\documentclass[DIV15, a4paper]{scrartcl}

\usepackage[ruled, linesnumbered, vlined, commentsnumbered]{algorithm2e}
\usepackage{authblk}
\usepackage{prettyref}
\usepackage{graphicx}
\usepackage{subfig}
\usepackage{booktabs}
\usepackage{colortbl}
\usepackage{amssymb}
\usepackage{amsfonts}
\usepackage{amsmath}
\usepackage{amsthm}
\usepackage{multirow}
\usepackage{tabularx}
\usepackage{footnote}
\usepackage{threeparttable}
\usepackage{hyperref}
\usepackage{amsopn}
\usepackage{hhline}
\usepackage{cite}
\usepackage{subfig}
\usepackage{setspace}

\usepackage{xcolor}  % Required for custom colors
\definecolor{myblue}{RGB}{34,31,217}

\definecolor{mycyan}{gray}{.7}

\newtheorem{theorem}{Theorem}
\newtheorem{proposition}{Proposition}
\newtheorem{corollary}{Corollary}
\newtheorem{definition}{Definition}

\newtheorem{lemma}{Lemma}

\DeclareMathOperator*{\argmin}{argmin}

\newrefformat{fig}{Fig.~\ref{#1}}
\newrefformat{tab}{Table~\ref{#1}}
\newrefformat{sec}{Section~\ref{#1}}
\newrefformat{app}{Appendix~\ref{#1}}
\newrefformat{alg}{Algorithm~\ref{#1}}
\newrefformat{property}{Property~\ref{#1}}
\newrefformat{theorem}{Theorem~\ref{#1}}
\newrefformat{corollary}{Corollary~\ref{#1}}
\newrefformat{proposition}{Proposition~\ref{#1}}
\newrefformat{def}{Definition~\ref{#1}}
\newrefformat{lemma}{Lemma~\ref{#1}}
\newrefformat{eq}{equation~(\ref{#1})}

\usepackage{graphicx}
\definecolor{Gray}{gray}{0.9}
\usepackage{scrpage2}

\usepackage{lscape}

\begin{document}

\title{\textbf\LARGE\fontfamily{cmss}\selectfont Dominance Move: A Measure of Comparing Solution Sets in Multiobjective Optimization}

%\author[1]{\normalsize\fontfamily{lmss}\selectfont Ke Li\footnote{Department of Computer Science, University of Exeter, EX4 4QF, UK. Email: k.li@exeter.ac.uk}\hspace{10mm} Kalyanmoy Deb\footnote{COIN Laboratory, Department of Electrical and Computer Engineering, Michigan State University, MI 48824, USA. Email: kdeb@egr.msu.edu}\hspace{10mm} Xin Yao\footnote{CERCIA, School of Computer Science, University of Birmingham, B15 2TT, UK. Email: x.yao@cs.bham.ac.uk}}

\author[1]{\normalsize\fontfamily{lmss}\selectfont Miqing Li and Xin Yao}
\affil[1]{\normalsize\fontfamily{lmss}\selectfont CERCIA, School of Computer Science, University of Birmingham, Birmingham B15 2TT, U.~K.}
\affil[$\ast$]{\normalsize\fontfamily{lmss}\selectfont Email: limitsing@gmail.com, x.yao@cs.bham.ac.uk}

\renewcommand\Authands{Miqing Li and Xin Yao}

\date{}
\maketitle

{\normalsize\fontfamily{lmss}\selectfont\textbf{Abstract:} One of the most common approaches for multiobjective optimization is to
		generate a solution set that well approximates the whole Pareto-optimal frontier to
		facilitate the later decision-making process.
		However,
		how to evaluate and compare the quality of different solution sets remains challenging.
		Existing measures typically require additional problem knowledge and information,
		such as a reference point or a substituted set of the Pareto-optimal frontier.
		In this paper,
		we propose a quality measure,
		called dominance move (DoM),
		to compare solution sets generated by multiobjective optimizers.
		Given two solution sets,
		DoM measures the minimum sum of move distances for one set to weakly Pareto dominate the other set.
		DoM can be seen as a natural reflection of the difference between two solutions,
		capturing all aspects of solution sets' quality,
		being compliant with Pareto dominance,
		and does not need any additional problem knowledge and parameters.
		We present an exact method to calculate the DoM in the biobjective case.
		We show the necessary condition of constructing the optimal partition for a solution set's minimum move,
		and accordingly propose an efficient algorithm to recursively calculate the DoM.
		Finally,
		DoM is evaluated on several groups of artificial and real test cases as well as
		by a comparison with two well-established quality measures.}
	
{\normalsize\fontfamily{lmss}\selectfont\textbf{Keywords:} multiple criteria, quality measure, metaheuristics}

\section{Introduction}

In multiobjective optimization,
it is often desirable for search algorithms to provide the decision maker (DM)
with a representative subset of the whole Pareto-optimal (efficient) frontier.
This largely avoids overload from both the computational and decision-making perspectives \cite{Sayin2005, Wallenius2008, Masin2008, Karasakal2009}.
Over the past several decades,
a variety of search techniques (called multiobjective optimizers here) have emerged ranging from exact methods
to heuristics or metaheuristics \cite{Miettinen1999, Deb2001, Ehrgott2006, Coello2007},
aiming at finding a ``good'' representation of the Pareto-optimal frontier.
However,
the meaning of ``good'' is ambiguous \cite{Faulkenberg2010}.
There is no clear definition in the operations research and optimization community of
what a good representation of the Pareto-optimal frontier should be.
This naturally leads to a question ---
how to evaluate and compare the quality of solution sets obtained by different multiobjective optimizers.

Quality evaluation of solution sets in the context of multiobjective optimization has gained wide attention
in both the operations research and computational intelligence areas \cite{Daniels1992, Hansen1998, Zitzler2003, Knowles2006, Faulkenberg2010}.
\cite{Daniels1992} considered the analytical evaluation of the solution quality of multiobjective heuristics,
and defined approximation errors as the value penalty incurred by approximating a Pareto-optimal solution with its heuristic alternative.
\cite{Sayin2000} proposed three metrics, coverage error, uniformity level and cardinality,
to evaluate the coverage of a solution set to the Pareto-optimal frontier,
the uniformity among solutions in the set
and the number of solutions in the set,
respectively.
To evaluate a solution set' quality in terms of both convergence and diversity,
\cite{Zitzler1999} measured the volume of the objective space
enclosed by the set and a reference point.
This is the well-known hypervolume metric.
\cite{Veldhuizen1998} considered the convergence of a solution set to the Pareto-optimal frontier
and defined generational distance (GD) by calculating the average distance from all solutions in the set to the optimal frontier.
Later,
GD was modified in reverse by considering the distance from the optimal frontier (represented by a reference set) to the solution set,
called inverted generational distance (IGD) \cite{Bosman2003}.
This measure can reflect both convergence and diversity of a solution set.
\cite{Carlyle2003} introduced the integrated preference functional (IPF)
to measure the quality of a solution set for biobjective optimization problems.
In their work,
the form of the decision maker's value function was represented as a convex combination of objectives,
and thus only supported points contribute to the IPF result.
Later,
\cite{Kim2006} presented an extension of the IPF for general $k$-objective optimization problems.
Recently,
\cite{Bozkurt2010} considered the weighted Tchebycheff function for the calculation of the IPF measure,
thus counting all nondominated points (both supported and unsupported) of a solution set.
On the other hand,
\cite{Zitzler2003} carried out theoretical study of quality measures.
They analysed theoretical limitations of unary quality measures
(i.e., the measure that assigns one solution set a numerical value that reflects a certain quality aspect),
and stated the strengths of binary quality measures
(i.e., the measure that directly uses a numerical value to describe how much better one solution set is than another).

Despite the activity of the design and analysis of quality evaluation techniques,
evaluating solution sets encounter various difficulties in practical applications \cite{Zitzler2008, Lizarraga2009, Bozkurt2010}.
Some measures evaluate only one aspect of a solution set's quality,
such as GD.
Many measures do not comply with the Pareto dominance relation of two solution sets \cite{Knowles2002}.
This happens a lot in those measures which solely evaluate the diversity of a solution set \cite{Li2014d}.
In addition,
the Pareto-optimal frontier (or its substituted set) of a given problem is often required
to compare solution sets,
such as in the coverage error, GD, IGD, and Daniels' approximation errors measure.
This,
however,
is commonly unavailable in practice.
Moreover,
some quality measures require additional problem knowledge in the evaluation of solution sets,
for example, the reference point in the hypervolume metric and the ideal point in the Tchebycheff-based IPF.
This may affect the evaluation results,
especially when comparing several solution sets with different spatial locations \cite{Lizarraga2008a}.
Finally,
parameter setting is an important issue in those parameter-dependent quality measures.
In such measures,
the accuracy of evaluation results depends largely on a proper choice of the parameter(s);
also,
the sensitiveness is often affected by the cardinality of a solution set and the dimension of its solution vectors.

In this paper,
we propose a new quality measure,
called dominance move (DoM),
to compare solution sets obtained by multiobjective optimizers.
DoM measures the minimum sum of move distances needed to make a set
(weakly) Pareto dominate another set.
It is able to capture all aspects of solution sets' quality,
i.e., the closeness to the Pareto-optimal frontier,
the extensity of the solution set covering,
the uniformity among solutions in the set,
and the number of nondominated solutions in the set.
Moreover,
DoM is Pareto compliant and does not need any additional problem knowledge and parameters.
Most importantly,
the proposed measure is of highly intuitive ---
it can be seen as a natural reflection of the differences between solution sets.

We present an exact calculation method for the DoM measure in the biobjective case.
We show the necessary condition of constructing the optimal partition for a solution set's minimum move,
and accordingly propose an efficient algorithm (with the computational complexity $O(N\log N)$
where $N$ is the cardinality of the set) to calculate the DoM result.

The rest of this paper is structured as follows.
In \S 2,
we give the main notations and terminology used and introduce some related work.
Section 3 is devoted to the description of the proposed DoM.
We present an exact method to calculate the DoM measure in \S 4.
Section 5 evaluates DoM,
including several artificial examples to develop the reader's intuition,
a comparison study with popular quality measures,
and two practical examples on combinatorial and continuous optimization problems.
Finally,
concluding remarks are provided in \S 6.

\section{Preliminaries and Related Work}

In this section,
first we briefly introduce several definitions that are used in the comparison between solutions and
accordingly between solution sets.
We then review some quality measures in the literature.

\subsection{Terminology}

Without lost of generality,
we consider a minimization problem with $m$ objective functions
$f^i : X \rightarrow \mathbb{R}^m$, $1\leq i \leq m$.
The objective functions map a point $\mathbf{x} \in X$ in the decision space to an objective vector
$f(\mathbf{x}) = (f^1(\mathbf{x}),...,f^m(\mathbf{x}))$ in the objective space $Z \subset \mathbb{R}^m$.
In view of that in most case only objective vectors are considered in quality evaluation,
we only define the comparison relation based on objective vectors.
In addition,
for reasons of simplicity we refer to an objective vector also as a \textit{solution}
(despite it being originally called in $X$)
and the outcome of a multiobjective optimizer as a \textit{solution set}.

Let two solutions $p, q \in Z$.
Solution $p$ is said to \textit{weakly dominate} $q$ (denoted as $p\preceq q$) if and only if $p^i \leq q^i$ for $1 \leq i \leq m$.
If there exists at least one objective $j$ on which $p^j < q^j$,
we say that $p$ \textit{dominates} $q$ (denoted as $p\prec q$).
A solution $p \in Z$ is called \textit{Pareto optimal} (efficient) if there is no other $q \in Z$ that dominates $p$.
The set of all Pareto optimal solutions of a multiobjective optimization problem is called its Pareto-optimal frontier.

The relation between solutions can be naturally extended to solution sets \cite{Zitzler2003}.
Let two solution sets $P, Q \subset Z$.
Solution set $P$ is said to \textit{weakly dominate} $Q$ (denoted as $P\preceq Q$)
if every solution $q \in Q$ is weakly dominated by at least one solution $p \in P$.
If for every solution $q \in Q$ there exists at least one solution $p \in P$ that dominates $q$,
we say that $P$ dominates $Q$ (denoted as $P\prec Q$).
We can see that the weak dominance relation between two sets does not rule out their equality,
while the dominance relation does completely.
There thus exists another situation that $P$ weakly dominates but does not equal $Q$.
That is,
every solution in $Q$ is weakly dominated by one solution in $P$
but there is at least one solution in $P$ that is not weakly dominated by any solution in $Q$
(i.e., $P\preceq Q \bigwedge Q\npreceq P$).
This relation represents the most general and weakest form of superiority between two solution sets
and was defined as $P$ being \textit{better} $Q$ (denoted as $P\vartriangleleft Q$) in \cite{Zitzler2003}.
Put it simply,
$P\vartriangleleft Q$ means that $P$ is at least as good as $Q$,
while $Q$ is not as good as $P$.

\subsection{Related Work}

With the development of effective search techniques in multiobjective optimization,
the issue of quality evaluation has become increasingly important.
Over the past several decades,
a large number of quality measures have been emerging not only
in the operations research \cite{Vira1983, Haimes2000} and evolutionary computation fields \cite{Fonseca1996, Okabe2003, Zitzler2008},
but also in other fields such as mechanical design \cite{Wu2001, Farhang-Mehr2003} and
software engineering \cite{Wang2016}.
%Next,
%we will review several widely-used quality measures in brief.

As the most general form of preference between solution sets in multiobjective optimization,
the Pareto dominance relation has naturally been considered in solution sets' quality comparison.
However,
in most cases this comparison is not feasible in practice.
On the one hand,
solution sets obtained by multiobjective optimizers are typically non-dominated to each other.
On the other hand,
people may be interested in more precise statements that quantify the difference between solution sets.

An alternative is to consider the dominance relation between solutions from different sets (rather than between sets).
For example,
\cite{Zitzler1998} compared two solution sets by the number of solutions in one set
that are not dominated by any solution in the other set.
\cite{Sayin2000} counted the number of nondominated solutions in the obtained set.
However,
this comparison cannot provide any information of solution sets' distribution
(e.g., the coverage of solution sets over the Pareto-optimal frontier).
In addition,
solutions between the sets become increasingly incomparable
as the number of objectives in optimization problems grows.

\cite{Sayin2000} also considered the distribution of a subset of the Pareto-optimal solutions,
introducing two quality measures,
coverage error and uniformity level.
The former measures the coverage of the set to the optimal frontier,
and the latter quantifies the distance between neighboring points in the set.
These measures,
however,
only work on the Pareto-optimal solutions and thus are infeasible for solution sets obtained by heuristics.
Similarly,
some popular quality measures in evolutionary multiobjective optimization,
such as GD \cite{Veldhuizen1998}, IGD \cite{Bosman2003} and their variants \cite{Schutze2012, Ishibuchi2015, Ishibuchi2015b},
require a reference set that well represents the Pareto-optimal frontier of the problem in their evaluation process.
However,
it is difficult to specify such a reference set without the knowledge of the Pareto optimal frontier.
Different reference sets can easily lead to inconsistent evaluation results between the solution sets \cite{Li2015b}.

\cite{Carlyle2003} introduced a quality measure based on the integrated preference functional (IPF),
without the requirement of the knowledge of the Pareto-optimal frontier.
This measure used partial information on the decision maker's value function
and was designed for biobjective optimization problems.
Later,
\cite{Kim2006} extended this IPF measure,
enabling it feasible for general $k$-objective optimization problems.
However,
since the the form of the decision maker's value function was represented as a convex combination of objectives,
only supported points in a solution set contribute to the IPF result.
To address this issue,
\cite{Bozkurt2010} considered the weighted Tchebycheff function as the value function in IPF,
which makes the evaluation result include the information of unsupported points.
This modification introduces an additional parameter,
the ideal point (for the calculation of the Tchebycheff function),
which may affect the evaluation result to some extent.

As a well-known measure for approximation algorithms in operations research and theory \cite{Helbig1994, Papadimitriou2000, Erlebach2002, Vaz2015} as well as evolutionary multiobjective optimization \cite{Laumanns2002, Bringmann2015},
the $\epsilon$-approximation can naturally be used to quantitatively compare solution sets.
\cite{Zitzler2003} presented an $\epsilon$-approximation based measure (called $\epsilon$ indicator)
with two versions,
additive $\epsilon$ indicator and multiplicative $\epsilon$ indicator.
Here,
we consider the additive $\epsilon$ indicator and similar results can be found in the multiplicative version.
For two solution sets $P$ and $Q$,
the additive $\epsilon$ indicator $I_{\epsilon}(P, Q)$ is the minimum value that can be added to each solution in $Q$
such that they become weakly dominated by at least one solution in $P$.
Formally,
the additive $\epsilon$ indicator is calculated as
%%%% equation 1 %%%%
\begin{equation}
I_{\epsilon}(P,Q) = \max_{q\in Q} ~\min_{p\in P} \max_{i\in \{1...m\}} p^i - q^i
\label{eq:dominancedistance}
\end{equation}
where $p^{i}$ denotes the objective value of solution $p$ in the $i$th objective
and $m$ is the number of objectives.

%%%% Fig. 1 %%%%
\begin{figure}[tb]
	\begin{center}
		\footnotesize
		\begin{tabular}{cc}
			\includegraphics[scale=0.4]{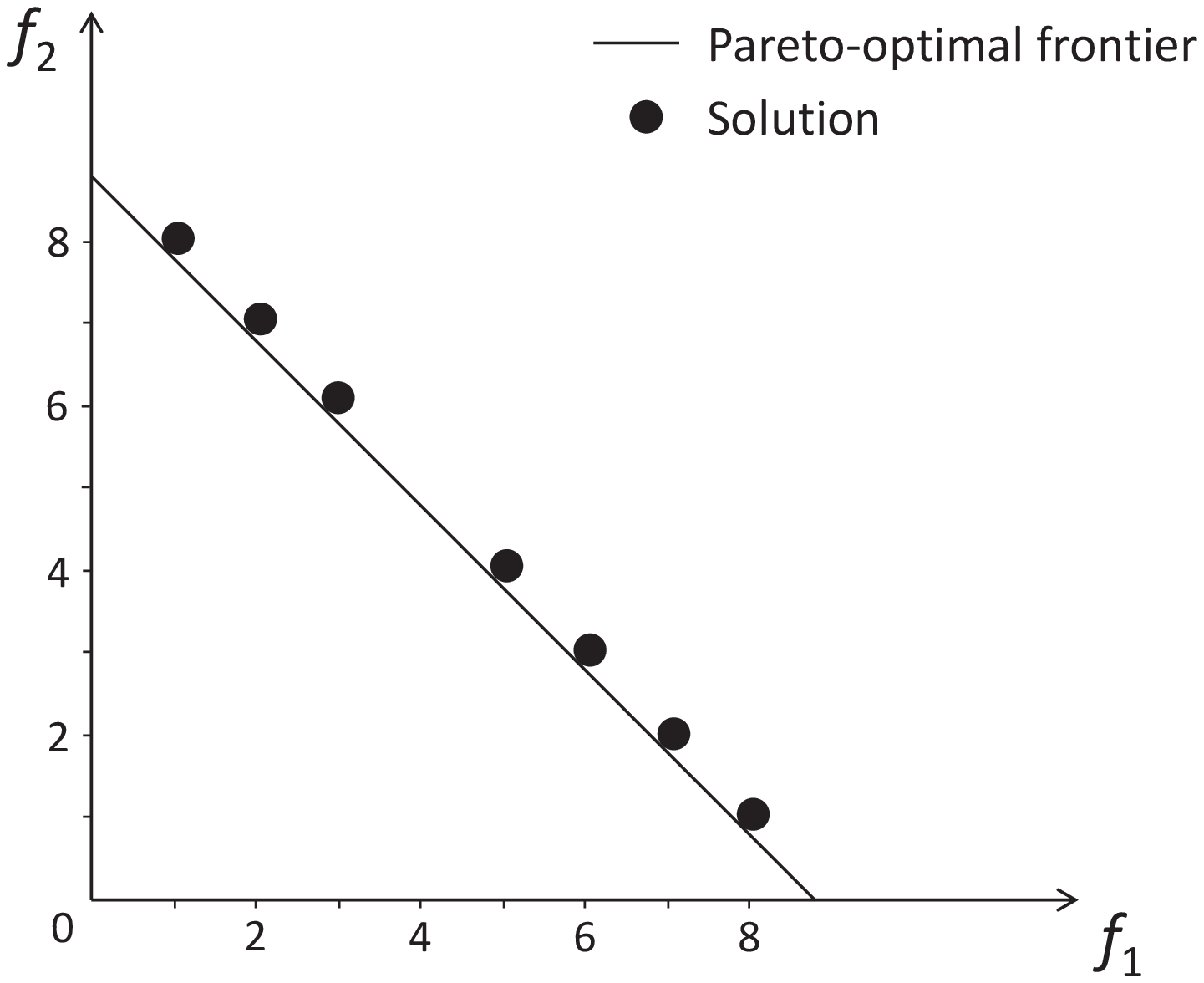} &
			\includegraphics[scale=0.4]{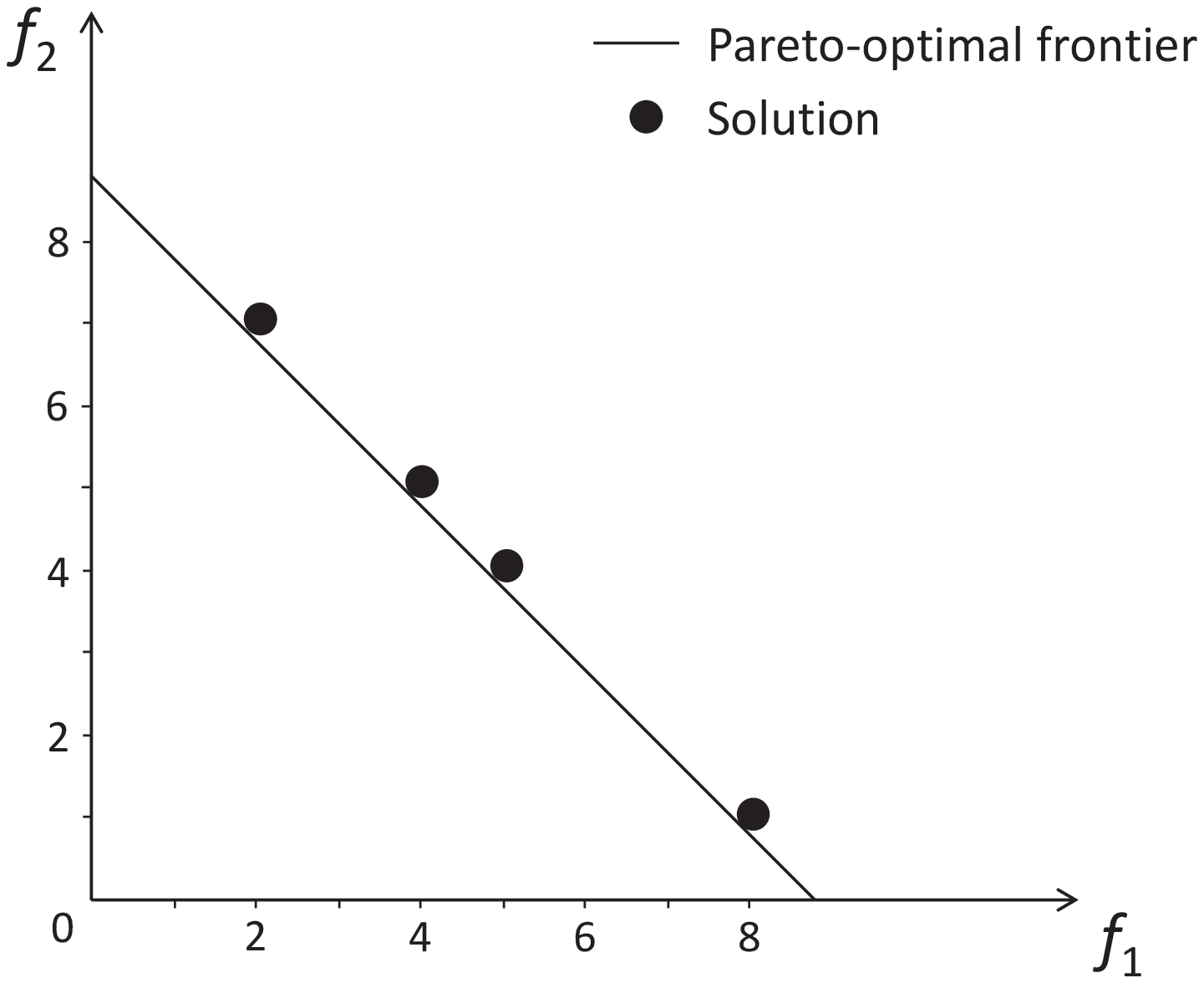} \\
			(a) Solution set $P$ & (b) Solution set $Q$ \\
		\end{tabular}
	\end{center}
	%\vspace{4mm}
	\caption{An example that the $\epsilon$ indicator makes an inaccurate evaluation of quality comparison between solution sets.
		Solution set $P$ has a better coverage over the Pareto-optimal frontier than solution set $Q$,
		but the two sets has the same comparison result (i.e., $I_{\epsilon}(P, Q) = I_{\epsilon}(Q, P) = 1$).}
	\label{Fig:Epsilon}
\end{figure}

As pointed out by \cite{Zitzler2003},
the $\epsilon$ indicator has some desirable features,
such as no need of a reference set,
complying with the Pareto dominance relation,
and representing natural extension to the evaluation of approximation schemes in operations research and theory.
However,
one weakness of the $\epsilon$ indicator is that its evaluation result is only related to
one particular solution in either solution set.
This could lead to an inaccurate evaluation of quality comparison between solution sets.
\mbox{Figure~\ref{Fig:Epsilon}} gives an example that the $\epsilon$ indicator fails to distinguish between solution sets ($P$ and $Q$).
As can be seen from the figure,
$P$ has more solutions and a better coverage over the Pareto-optimal frontier than $Q$,
but the two sets have the same comparison result ($I_{\epsilon}(P, Q) = I_{\epsilon}(Q, P) = 1$).

In addition,
the $\epsilon$ indicator only considers one particular objective in comparing the difference between two solutions
(i.e., the objective on which one solution performs worst relative to another solution).
This ignores the difference on other objectives
and naturally leads to an information loss.
This information loss becomes more severe as the number of optimization problems' objective increases.
Consider two 10-objective solutions,
$p=(0,0,0,...,0,1)$ and $q=(1,1,1,...,1,0)$.
Solution $p$ performs better on nine objectives and solution $q$ better only on the last objective,
but they have same comparison result ($I_{\epsilon}(p, q) = I_{\epsilon}(q, p) = 1$).

%%%% Fig. 2 %%%%
\begin{figure}[tb]
	\begin{center}
		\footnotesize
		\begin{tabular}{cc}
			\includegraphics[scale=0.4]{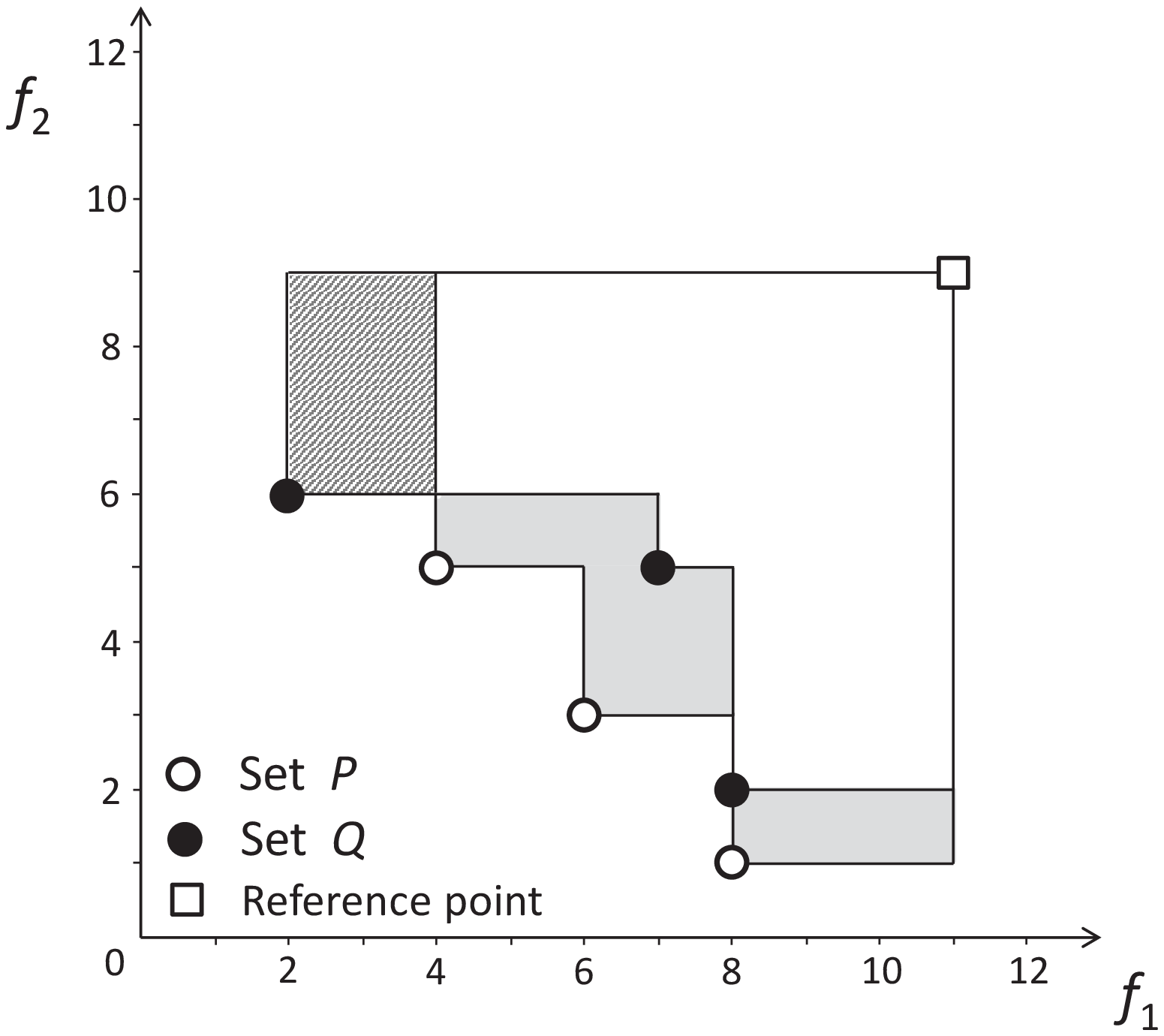} &
			\includegraphics[scale=0.4]{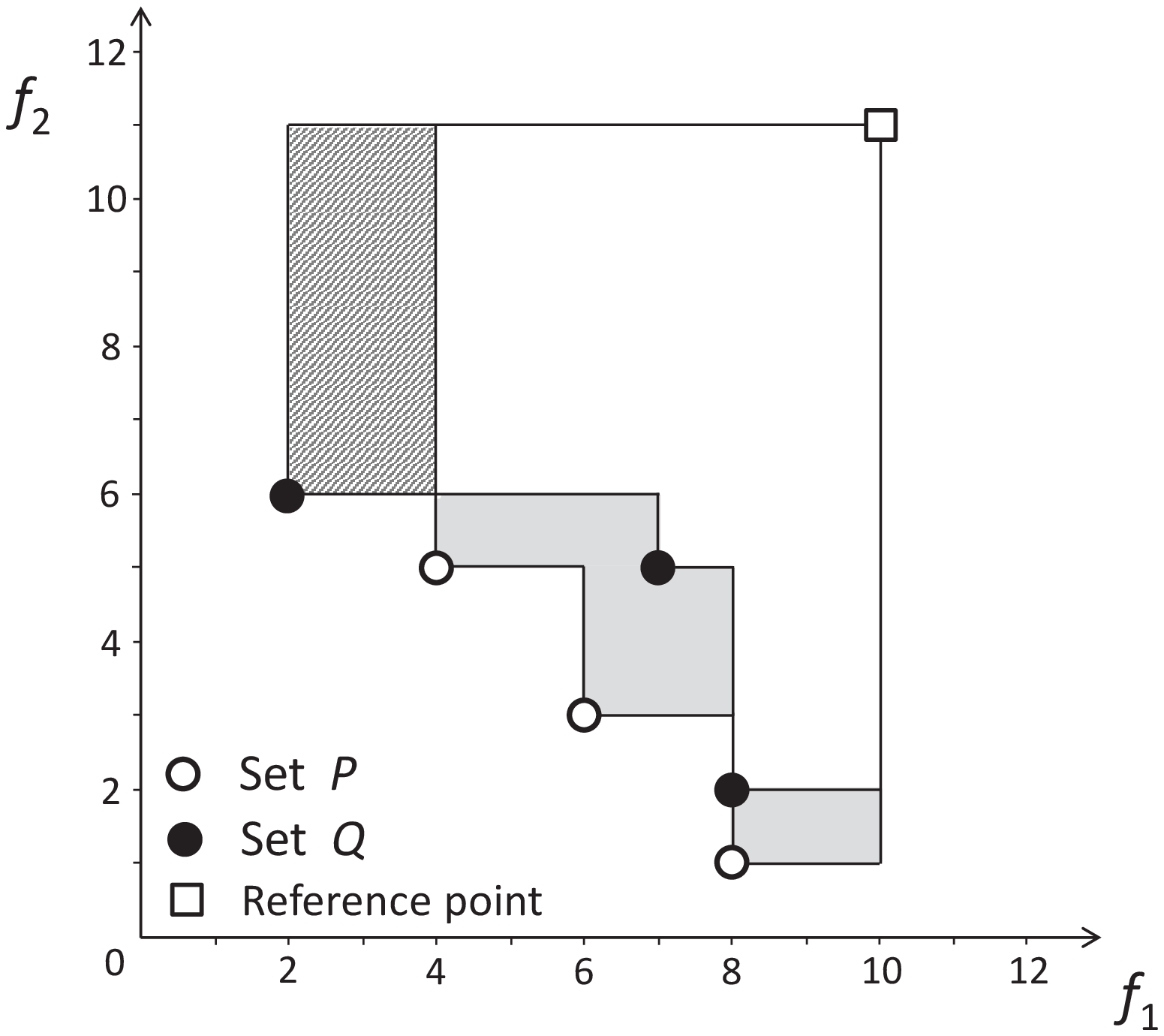} \\
			(a) With reference point $(11,9)$ & (b) With reference point $(10,11)$ \\
		\end{tabular}
	\end{center}
	%\vspace{4mm}
	\caption{An example that the HV metric makes an inconsistent evaluation of solution sets under the different reference points,
		where grey area $\subset \textrm{HV}(P)$ but $\nsubseteq \textrm{HV}(Q)$,
		and hatched area $\subset \textrm{HV}(Q)$ but $\nsubseteq \textrm{HV}(P)$.
		With reference point $(11,9)$, $\textrm{HV}(P) = 44 > \textrm{HV}(Q) = 40$,
		while with reference point $(10,11)$, $\textrm{HV}(P) = 48 < \textrm{HV}(Q) = 49$.}
	\label{Fig:HV}
\end{figure}

The hypervolume (HV) metric \cite{Zitzler1999} is one of the most popular quality measures in multiobjective optimization.
It calculates the volume of the space enclosed by a solution set and a reference point,
and a large value is preferable.
The HV of a solution set $P$ can be described as the Lebesgue measure $\Lambda$ of the union hypercubes $h(p,r)$
defined by $p=(p^1,...,p^m) \in P$ and the reference point $r=(r^1,...,r^m)$:
%%%% equation 2 %%%%
\begin{equation}
\textrm{HV} = \Lambda\left(\bigcup_{p\in P}h(p,r)\right)
\end{equation}
where $h(p,r) = [p^1, r^1] \times ... \times [p^m, r^m]$ ($p^i \leq r^i$ for all $i$).

The HV indicator has good theoretical properties \cite{Zitzler2003} and can give
a comprehensive evaluation of a solution set in terms of both convergence and diversity.
While the computational complexity of calculating HV increases exponentially with the number of objectives,
the Monte Carlo sampling method can provide a good balance between accuracy and running time \cite{Bader2011,Bringmann2013}.
However,
the HV indicator is sensitive to the choice of the reference point.
How to choose a proper reference point is not trivial and
different reference points can lead to inconsistent evaluation results \cite{Knowles2002}.
Take two solution sets ($P$ and $Q$) in \mbox{Figure~\ref{Fig:HV}} as an example.
When the reference point is set to $(11,9)$ (\mbox{Figure~\ref{Fig:HV}(a)}),
$\textrm{HV}(P) = 44 > \textrm{HV}(Q) = 40$.
When the reference point is $(10,11)$ (\mbox{Figure~\ref{Fig:HV}(b)}),
$\textrm{HV}(P) = 48 < \textrm{HV}(Q) = 49$.

%Is it better to add some comments about the proposed measure solving all of the above problems?

\section{The Proposed Measure}

Dominance move (DoM) is a measure of comparing two sets of multi-dimensional points (i.e., vectors).
It considers the move of points in one set to make this set weakly dominate the other set.
DoM can be defined as follows.

\begin{definition}
	Let $P$ be a set of points $\{p_1, p_2,...,p_{n}\}$ and $Q$ be a set of points $\{q_1, q_2,...,q_{l}\}$.
	The dominance move of $P$ to $Q$ (denoted as $D(P,Q)$) is the minimum total distance of moving points of $P$
	such that any point in $Q$ is weakly dominated by at least one point in $P$.
	That is,
	we move $p_1, p_2,...,p_{n}$ to positions $p'_1, p'_2,...,p'_n$\footnote{If $p_i$ keeps still, we can regard it as $p_i = p'_i$.} (thus constituting $P'$) such that
	1) $P'$ weakly dominates $Q$ and
	2) the total of the move from $p_1, p_2,...,p_n$ to $p'_1, p'_2,...,p'_n$ is minimum.
\end{definition}

Mathematically,
DoM can be expressed as
%%%% equation 3 %%%%
\begin{equation}
D(P,Q) = \min_{P'\preceq Q} ~\sum_{i=1}^{n} d(p_i, p'_i)
\end{equation}
%%%% equation 4 %%%%
\begin{equation}
d(p_i, p'_i) = \sum_{j=1}^{m} |p_i^j - p_i'^{j}|
\label{eq:DoMset}
\end{equation}
where $P=\{p_1,...,p_n\}$, $P'=\{p'_1,...,p'_n\}$,
$p_i^{j}$ denotes the value of solution $p_i$ in the $j$th objective,
and $m$ is the number of objectives.

\begin{figure}[tb]
	\begin{center}
		\footnotesize
		\begin{tabular}{@{}c@{}c@{}c@{}}
			\includegraphics[scale=0.35]{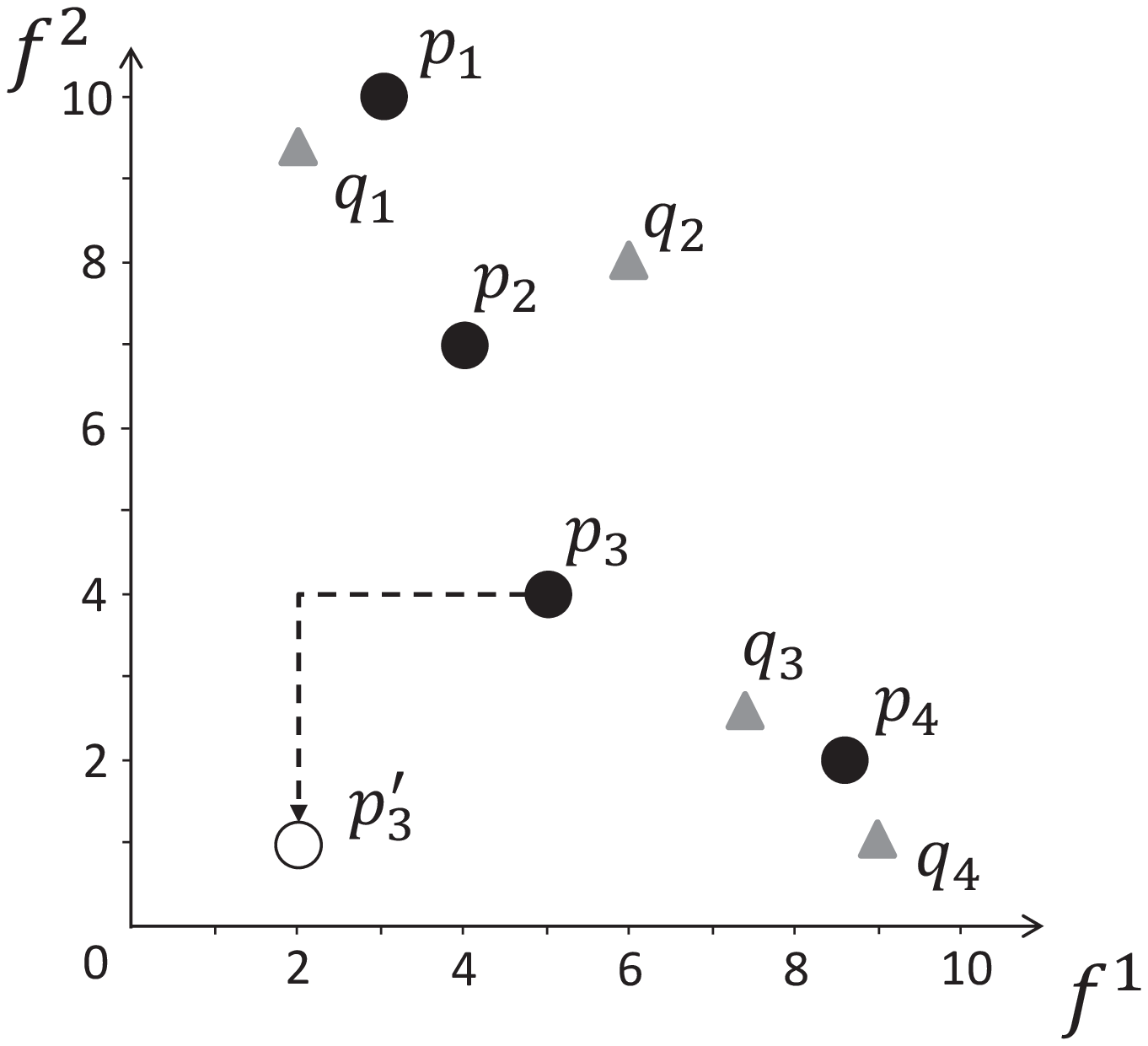}~~ & ~~
			\includegraphics[scale=0.35]{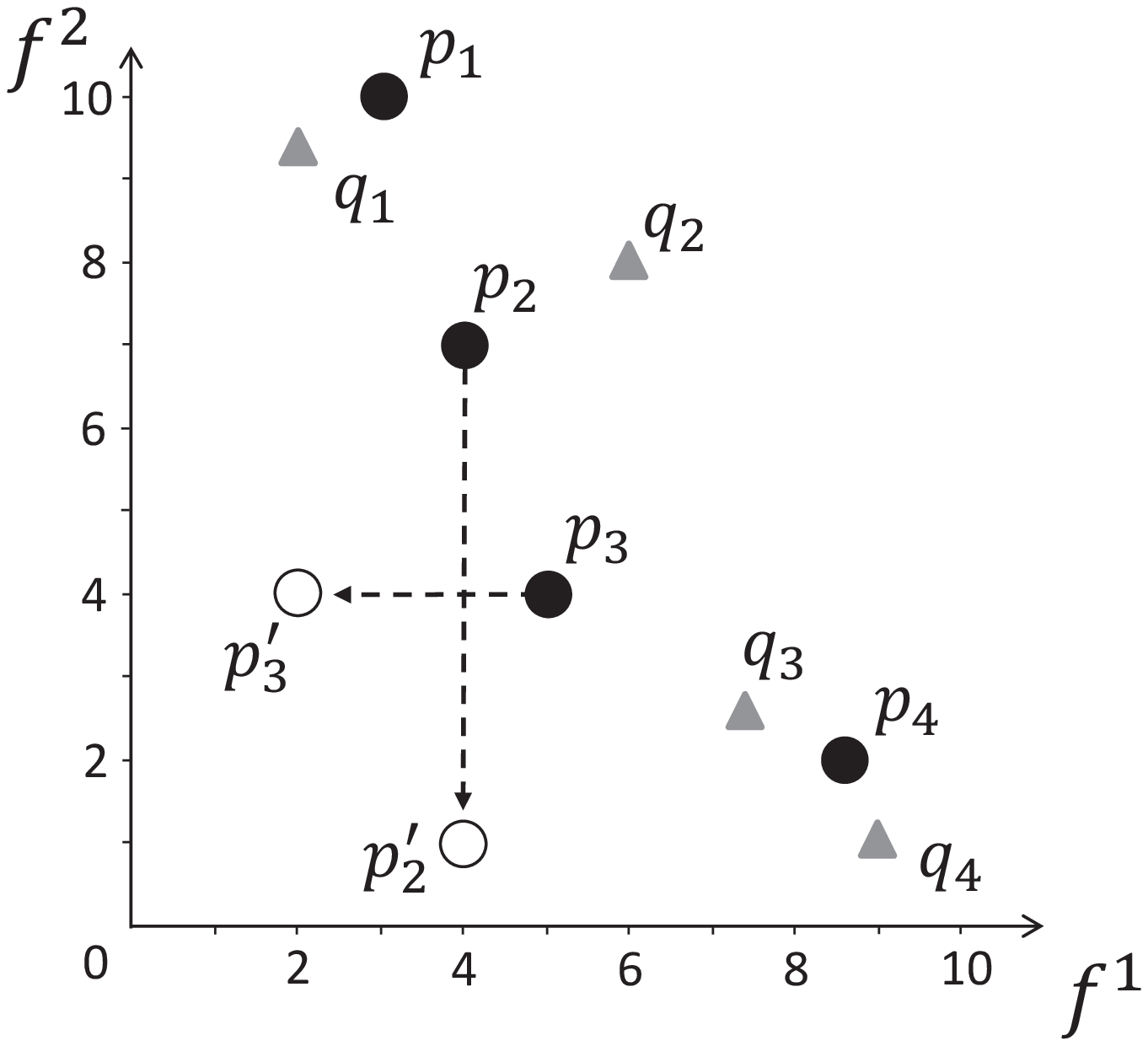}~~ & ~~
			\includegraphics[scale=0.35]{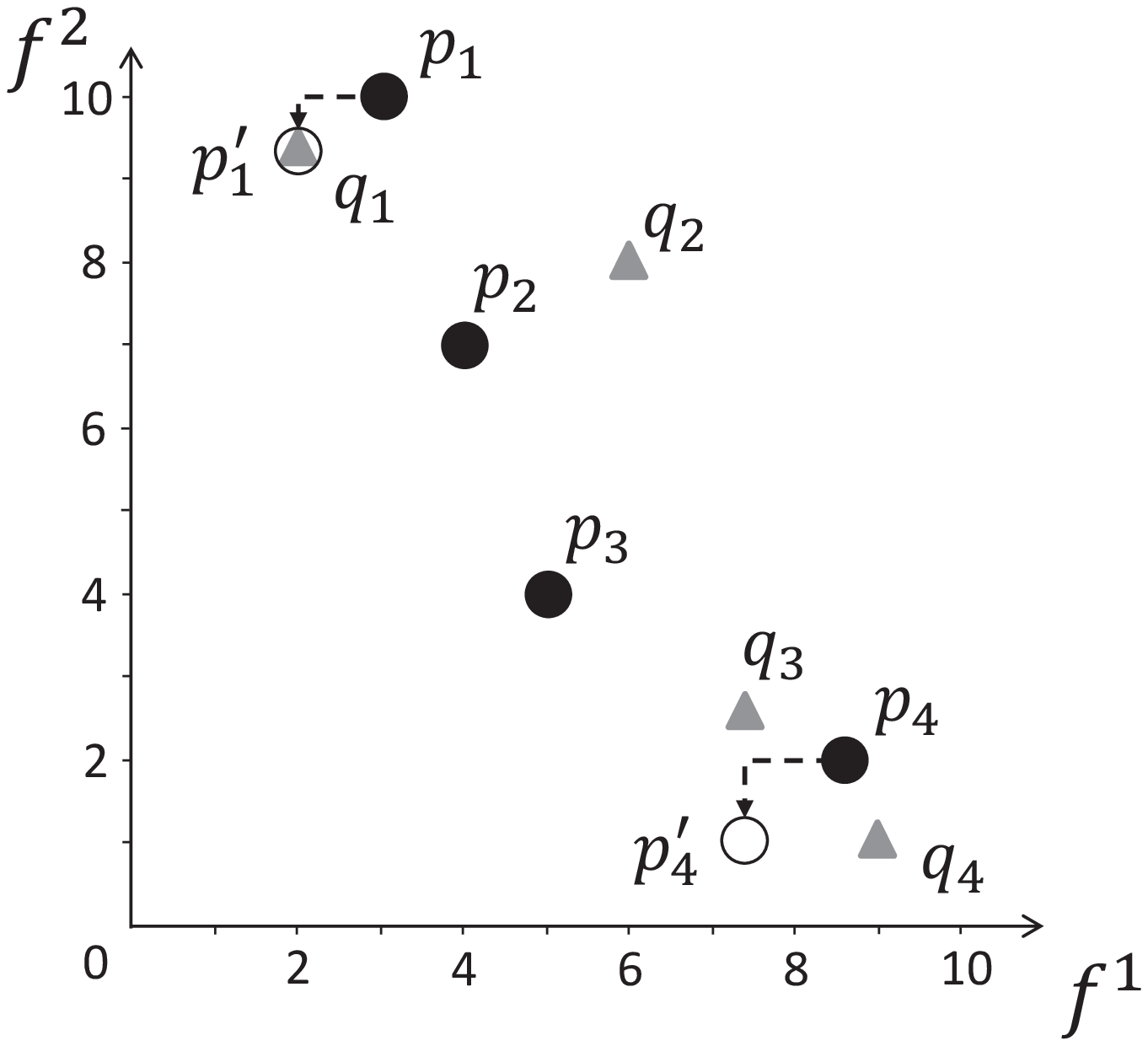} \\
			(a) Total move of $P$ is & (b) Total move of $P$ is & (c) Total move of $P$ is \\
		\end{tabular}
	\end{center}
	%\vspace{4mm}
	\caption{Different ways of moving points of $P = \{p_1, p_2, p_3, p_4\}$ to cover (weakly dominate) $Q= \{q_1, q_2, q_3, q_4\}$.
		(a) $p_3$ moving to $p_3'$ to cover all four points of $Q$.
		(b) $p_2$ and $p_3$ moving to $p_2'$ and $p_3'$ to cover $\{q_2, q_3, q_4\}$ and $\{q_1, q_2\}$,
		respectively.
		(c) $p_1$ and $p_4$ moving to $p_1'$ and $p_4'$ to cover $\{q_1\}$ and $\{q_3, q_4\}$,
		respectively, as $q_2$ is already covered by one point in $P$.}
	\label{Fig:DoMexample}
\end{figure}

Apparently,
to make $P'\preceq Q$,
any point $p\in P$ either stays still or moves to a ``better'' position (i.e., $p' \preceq p$).
There are numerous ways to make this move.
For example,
we could move only one point of $P$ to make it cover (weakly dominate) all points in $Q$;
we could also move some points of $P$ to make them together cover all points in $Q$;
or we could directly consider the minimum move of $P$'s points to each point of $Q$.
\mbox{Figure~\ref{Fig:DoMexample}} gives three examples of this move.
In \mbox{Figure~\ref{Fig:DoMexample}(a)},
$p_3$ moves to the $p_3'$ position to cover all four points of $Q$.
In \mbox{Figure~\ref{Fig:DoMexample}(b)},
$p_2$ and $p_3$ move to $p_2'$ and $p_3'$ to cover $\{q_2, q_3, q_4\}$ and $\{q_1, q_2\}$,
respectively.
In \mbox{Figure~\ref{Fig:DoMexample}(c)},
$p_1$ and $p_4$ move to $p_1'$ and $p_4'$ to cover $\{q_1\}$ and $\{q_3, q_4\}$,
respectively,
in view of that $q_2$ is already covered by one point of $P$.

%%%% Fig. 3 %%%%

Out of numerous possibilities of the above move,
the dominance move corresponds to the one that has the minimum distance.
This minimum distance reflects how far one solution set ($P$) needs to move to cover another solution set ($Q$).
In other words,
it indicates performance inferiority of $P$ to $Q$ (i.e., the advantage of $Q$ over $P$).
It is clear that $D(P,Q)$ is always larger than or equal to $0$.
A small value indicates that $P$ performs well relative to $Q$.
This implies the quality in both convergence and diversity.
A small $D(P,Q)$ means that for each point of $Q$ (whatever its location),
there exists at least one point of $P$ that only needs a little move at most to cover it.
Take \mbox{Figure~\ref{Fig:DoMexample}} as an example,
where the dominance move $D(P,Q)$ corresponds to the case of \mbox{Figure~\ref{Fig:DoMexample}(c)}.
As can be seen,
$P$ has a good coverage over $Q$ since only a little move of its points is needed to (weakly) dominate any point of $Q$.
In addition,
if $P\preceq Q$ then $D(P,Q) = 0$.

It is worth noting that the well-known $\epsilon$ indicator also measures the minimum value
added to one solution set to make it be weakly dominated by another set.
However,
it only considers one particular solution in either solution set.
This inevitably leads to information loss of the comparison between the two sets.
For the example in \mbox{Figure~\ref{Fig:DoMexample}},
the $\epsilon$ indicator ($I_{\epsilon}(P, Q)$) would only measure the advantage of $q_3$ over $p_4$,
regardless of the difference between the remaining solutions of the two sets.
A comparative test will be carried out later (Section 5.2) to demonstrate the difference
between the proposed DoM and the $\epsilon$ indicator (as well as other quality measures).
In the following,
we present several properties of DoM which can help further understand its behavior.

\begin{proposition}
	Let solution sets $P, Q, A, B, C \subset Z$ and points $p, q \in Z$.
	Then the following facts are true:
	
	(a) $P = Q \iff D(P,Q) = D(Q,P) = 0$.
	
	(b) $P \vartriangleleft Q$ (i.e., $P\preceq Q \bigwedge Q\npreceq P$, see Section~2.1) $\iff$ $D(P,Q) = 0 \bigwedge D(Q,P) > 0$.
	
	(c) $D(P,Q) \geq D(P\bigcup p, Q)$ and $D(P,Q) \leq D(P, Q\bigcup q)$.
	
	(d) Let $p \in P$ and $q \in Q$.
	If $\exists p' \in P, p'\prec p$,
	then $D(P,Q) = D(P/p, Q)$.
	Also, if $\exists q' \in Q, q'\prec q$ or $\exists p' \in P, p'\prec q$,
	then $D(P,Q) = D(P, Q/q)$.
	
	(e) If $A \preceq B$, then $D(A,C) \leq D(B,C)$ and $D(C,B) \leq D(C,A)$.
	
	(f) $D(A,B) + D(B,C) \geq D(A,B\bigcup C)$ and $D(A,B) + D(A,C) \geq D(A, B\bigcup C)$.
	\label{pro1}
\end{proposition}
\proof {Proof.}
Facts (a)--(d) follow directly from the definition of the dominance move.

(e) By the definition of $D(B,C)$,
for any point $c \in C$,
there exists one point $b \in B$ moving to $b'$ to weakly dominate $c$ in $D(B,C)$.
Since $A \preceq B$,
for point $b$ there exists one point $a \in A$ which weakly dominates $b$.
This means that the value of any objective of point $a$ is either equal to or smaller than that of point $b$.
Therefore,
the move distance from $a$ to $b'$ is equal to or smaller than that from $b$ to $b'$.
Given this and $b'\preceq c$,
we can construct a move for $A$ to weakly dominate $C$ whose total distance is equal to or smaller than $D(B,C)$.
In addition,
by the definition of the dominance move,
this distance is greater than or equal to $D(A,C)$.
Thus $D(A,C) \leq D(B,C)$.

The second inequality is proven analogously.

(f) Let $A_B$ be the new position to which $A$ moves in $D(A,B)$.
Then $A_B \preceq B$.
By Fact (e) above,
$D(A_B, C) \leq D(B, C)$ follows.
Let $A_{BC}$ be the position to which $A_B$ moves in $D(A_B, C)$.
Now,
we can obtain a path from $A$ to $A_B$ and then from $A_B$ to $A_{BC}$ (for $A$ to weakly dominate $B \bigcup C$),
whose move distance is equal to or smaller than $D(A,B) + D(B,C)$.
Likewise,
by the dominance move definition,
this distance is greater than or equal to $D(A, B \bigcup C)$.
Thus $D(A,B) + D(B,C) \geq D(A,B\bigcup C)$.

The second inequality is proven analogously.
%\Halmos
\endproof

Fact (b) implies a desirable property of the proposed measure in comparing two solution sets:
whenever one solution set $P$ is better than another set $Q$,
then $D(P, Q) < D(Q, P)$.
Fact (e) implies the relation of two solution sets when they are individually compared with a third set.
That is,
if $D(P_1, Q) < D(P_2, Q)$,
then $P_1$ will not be (weakly) dominated by $P_2$;
if $D(P, Q_1) < D(P, Q_2)$,
then $Q_2$ will not be (weakly) dominated by $Q_1$.

When considering the move of $P$ to cover $Q$,
any $q \in Q$ will be associated with a point $p \in P$ (i.e., $p$ moving to somewhere to weakly dominate it).
This is like a partition in the sense that
the set $Q = \{q_1, q_2,...,q_{l}\}$ is partitioned by points of $P$ into some groups\footnote{In principle,
	one point of $Q$ could be put into more than one group.
	However, this will naturally lead to a longer distance for $P$ to move.
	So here for brevity we only consider one point of $Q$ being in one group.} (subsets)
such that the union of these groups is $Q$ and each group corresponds to only one point of $P$.
For example,
in \mbox{Figure~\ref{Fig:DoMexample}(a)},
all points of $Q$ are put into one group corresponding to $p_3$.
In \mbox{Figure~\ref{Fig:DoMexample}(b)},
($q_1, q_2$) and ($q_3, q_4$) can be seen to be put into two groups corresponding to $p_3$ and $p_2$,
respectively.
In \mbox{Figure~\ref{Fig:DoMexample}(c)},
($q_1$), ($q_2$) and ($q_3, q_4$) can be seen into three groups corresponding to $p_1$, $p_2$ and $p_4$,
respectively.

After a partition is created,
for one group (denoted as $Q_s=\{q_{s1},...q_{sk}\}$),
its corresponding point $p$ needs to move to somewhere to weakly dominate all points in $Q_s$.
The minimum distance of this move is equal to the distance from $p$ to the ideal
point\footnote{The ideal point of a set of points is constructed by the best value of each objective for all points in the set.}
of $Q_s$ and $p$ (denoted as $I_{Q_s \cup p}$).
Formally,
it can be expressed as
%%%% equation 5 %%%%
\begin{equation}
d(p,Q_s) = \sum_{j=1}^m \left(p^j-\min\{p^j, q_{s1}^j, q_{s2}^j,...,q_{sk}^j\}\right)
\label{eq:DoMpoint}
\end{equation}
where $Q_s=\{q_{s1}, q_{s2},...,q_{sk}\}$,
$p^j$ denotes the objective value of point $p$ in the $j$th objective,
and $m$ is the number of objectives.

In fact,
$d(p,Q)$ can also be seen as the dominance move distance of one point ($p$) to a set ($Q$);
i.e., when $P = \{p\}$,
$D(P,Q) = D(p,Q) = d(p,Q)$.
Additionally,
when $Q = \{q\}$ (i.e., $\mid Q\mid = 1$),
$D(p,q)$ means the dominance move of point $p$ to point $q$,
namely, $D(p,q) = \sum_{j=1}^m (p^j-\min\{p^j, q^j\})$.
Note that in the calculation of the dominance distance (\mbox{Equations~(\ref{eq:DoMset}) and (\ref{eq:DoMpoint})}),
the move of the points is based on the Manhattan distance ($\ell_1$ norm).
In principle,
other distance metrics,
such as Euclidean distance,
could also be used.
We here consider the Manhattan distance in views of its desirable properties.
On the basis of the measure of the Manhattan distance,
when adding one point $q\in Q$ to an existing group $Q_s$ associated with $p$,
the additional move can be calculated by a direct comparison between $q$ and $I_{Q_s\cup p}$.
This indicates that we do not need to consider the order of points of $Q$ entering a group.
In other words,
for a group $Q_s$,
a point $p\in P$ can successively ``access'' (i.e., move to somewhere to cover) the points of $Q_s$ in any order,
and the total move distance is always the same
(namely, the distance of the move from $p$ to $I_{Q_s\cup p}$ directly).

The following proposition gives the connection with respect to the dominance move
when adding a set of points $Q_t$ into an existing group $Q_s$.

\begin{proposition}
	Let $P$ and $Q$ be two solution sets,
	$Q_s = \{q_{s1}, q_{s2},...,q_{sk}\} \subset Q$
	and $Q_t = \{q_{t1}, q_{t2},...,q_{ti}\} \subset Q$.
	Then the following equality
	%(with respect to the dominance move of one point to one set)
	holds for any point $p\in P$:
	%%%% equation 6 %%%%
	\begin{equation}
	d(p, Q_s\cup Q_t) = d(p, Q_s) + d(I_{Q_s\cup p},Q_t)
	\label{eq:DoMgroup}
	\end{equation}
	\label{pro2}
	where $I_{Q_s\cup p}$ denotes the ideal point of set $\{q_{s1}, q_{s2},...,q_{sk}, p\}$.
\end{proposition}

\proof {Proof.}
By \mbox{Equation~(\ref{eq:DoMpoint})},
we have
%%%% equation 7 %%%%
\[
\begin{split}
d(p, Q_s\cup Q_t) =~& \sum_{j=1}^m \left(p^j- \min\{p^j,q_{s1}^j,...,q_{sk}^j,q_{t1}^j,...,q_{ti}^j\} \right) \\
=~& \sum_{j=1}^m \left(p^j - \min\{p^j,q_{s1}^j,...,q_{sk}^j,q_{t1}^j,...,q_{ti}^j\} + \min\{p^j, q_{s1}^j,...,q_{sk}^j\} - \min\{p^j, q_{s1}^j,...,q_{sk}^j\} \right) \\
=~& \sum_{j=1}^m \left(p^j - \min\{p^j, q_{s1}^j,...,q_{sk}^j\} \right) + \sum_{j=1}^m \left(\min\{p^j, q_{s1}^j,...,q_{sk}^j\} - \min\{p^j,q_{s1}^j,...,q_{sk}^j,q_{t1}^j,...,q_{ti}^j\} \right)\\
=~& \sum_{j=1}^m \left(p^j - \min\{p^j, q_{s1}^j,...,q_{sk}^j\} \right) + \sum_{j=1}^m \left(I^j_{\{p,q_{s1},...,q_{sk}\}} - \min\{I^j_{\{p,q_{s1},...,q_{sk}\}}, q_{t1}^j,...,q_{ti}^j\} \right)\\
=~& d(p, Q_s) + d(I_{Q_s\cup p},Q_t)\\
\end{split}
\label{eq:pro1Proof}
\]
where $p^j$ denotes the objective value of point $p$ in the $j$th objective,
and $m$ is the number of objectives.
%\Halmos
\endproof

Since $I_{Q_s\cup p} \preceq p$,
by Fact (e) of \mbox{Proposition~\ref{pro1}},
it follows that $d(I_{Q_s\cup p},Q_t) \leq d(p,Q_t)$.
Substituting this in \mbox{Equation~(\ref{eq:DoMgroup})},
we have
\begin{corollary}
	$d(p, Q_s\cup Q_t) \leq d(p, Q_s) + d(p, Q_t)$
	\label{cor1}
\end{corollary}

The above results have presented properties of the dominance move of one point of $P$ to the groups which $Q$ is partitioned into.
%This allows us to redefine the dominance move of $P$ to $Q$ from the perspective of the partitioning.
Next,
we redefine the dominance move of $P$ to $Q$ from the perspective of the partitioning.

\begin{definition}
	Let solution sets $P,Q \subset Z$.
	We call $\mathcal{S}(P,Q) = (p_{s1}, Q_{s1}), (p_{s2}, Q_{s2}),...,(p_{sn}, Q_{sn})$ to be a partition of $P$ to $Q$
	if $Q_{s1} \bigcup Q_{s2} \bigcup ... \bigcup Q_{sn} = Q$,
	where $p_{s1}, p_{s2},..., p_{sn} \in P$ and $Q_{s1}, Q_{s2},..., Q_{sn} \subset Q$.
	Then the dominance move of $P$ to $Q$ corresponds to the partition satisfying that
	$\sum_{i=1}^n d(p_{si},Q_{si})$ is minimum.
	We call this partition an optimal partition of $P$ to $Q$.
	%	(denoted as ($\mathcal{S^*}(P,Q)$)).
\end{definition}

Note that there can be more than one partition satisfying that $\sum_{i=1}^n d(p_{si},Q_{si})$ is minimum.
So,
we may have several optimal partitions of $P$ to $Q$.
From the above definition,
the proposed measure $D(P,Q)$ is transformed into a problem of finding an optimal partition of $P$ to $Q$.
This partitioning problem has an important property below.

%The following theorem shows an important property with respect to such a partition.
%Note that since any point of $Q$ that is dominated by one point in $P$ will contribute nothing to the dominance move distance,
%for brevity we assume no point of $Q$ being dominated by points in $P$.

\begin{theorem}
	Let $\mathcal{S}(P,Q)$ be a partition of one set $P$ to another set $Q$.
	Let $\overline{Q}$ be the set of $Q$'s points of one or several groups in $\mathcal{S}$.
	Now if we can find a partition of $P$ to $\overline{Q}$ with a smaller dominance move than the original group(s) in $\mathcal{S}$,
	then we can construct a smaller partition of $P$ to $Q$ than $\mathcal{S}$.
	\label{the1}
\end{theorem}

\proof {Proof.}
Let the original partition of $P$ to $\overline{Q}$ in $\mathcal{S}$ be $(p_{s1},Q_{s1}), (p_{s2},Q_{s2}),..., (p_{si},Q_{si})$,
and the new smaller partition of $P$ to $\overline{Q}$ be $(p_{t1},Q'_{s1}), (p_{t2},Q'_{s2}),..., (p_{tj},Q'_{sj})$.
Then $\overline{Q} = Q_{s1} \bigcup Q_{s2} \bigcup...\bigcup Q_{si} = Q'_{s1} \bigcup Q'_{s2} \bigcup...\bigcup Q'_{sj}$
and $d(p_{s1},Q_{s1}) + d(p_{s2},Q_{s2})+...+ d(p_{si},Q_{si}) > d(p_{t1},Q'_{s1}) + d(p_{t2},Q'_{s2})+...+ d(p_{tj},Q'_{sj})$.
Let $Q_{t1}, Q_{t2},..., Q_{tj}$ be the original groups associated with $p_{t1}, p_{tj},..., p_{tj}$ in $\mathcal{S}$.
Now we have that
\[
\begin{split}
~& d(p_{s1},Q_{s1}) + d(p_{s2},Q_{s2})+...+ d(p_{si},Q_{si}) + d(p_{t1},Q_{t1}) + d(p_{t2},Q_{t2})+...+ d(p_{tj},Q_{tj}) \\
> ~& d(p_{t1},Q'_{s1}) + d(p_{t2},Q'_{s2})+...+ d(p_{tj},Q'_{sj}) + d(p_{t1},Q_{t1}) + d(p_{t2},Q_{t2})+...+ d(p_{tj},Q_{tj}) \\
\geq ~& d(p_{t1},Q'_{s1}\bigcup Q_{t1}) + d(p_{t2},Q'_{s2} \bigcup Q_{t2})+...+ d(p_{tj},Q'_{sj}\bigcup Q_{tj})
\end{split}
\]
where the last inequality is obtained by \mbox{Corollary~\ref{cor1}}.
This means that the points of the groups $Q_{s1}, Q_{s2},..., Q_{si}, Q_{t1}, Q_{t2},..., Q_{tj}$ in $\mathcal{S}$
can be repartitioned by $p_{t1}, p_{t2},...,p_{tj}$ with a smaller dominance move
(since $Q_{s1} \bigcup Q_{s2} \bigcup...\bigcup Q_{si} = Q'_{s1} \bigcup Q'_{s2} \bigcup...\bigcup Q'_{sj}$).
This will lead to a new partition of $P$ to $Q$ with a smaller dominance move
as long as we keep the other groups in the original $\mathcal{S}$ unchanged.
%\Halmos
\endproof

This theorem indicates that if merging or splitting some groups can lead to a smaller partition of them,
then these groups do not exist in any optimal partition.
This further results in several properties of the optimal partitioning.

%Next,
%we present several properties with respect to the optimal partition.

\begin{corollary}
	Let $\mathcal{S^*}(P,Q)$ be an optimal partition of a solution set $P$ to another set $Q$,
	and $Q_s \subset Q$ ($Q_s$ associated with $p_s\in P$) be one of the groups obtained by this partition.
	Then the following facts are true:
	
	(a) If $Q_{s} = \{q_s\}$ (i.e., $\mid Q_{s}\mid = 1$),
	then $p_s \in \{\argmin_{p\in P} d(p,q_s)\}$.
	
	(b) Considering the optimal partitioning of $P$ to $Q_s$,
	keeping $Q_s$ with $p_s$ corresponds to an optimal partition of them.
	
	(c) When $\mid Q_{s}\mid > 1$,
	let $q_{s1}, q_{s2}$ be two points in $Q_s$.
	In $Q$ there may exist some points ``between'' $q_{s1}$ and $q_{s2}$ (i.e., being weakly dominated by $I_{\{q_{s1}, q_{s2}\}}$).
	Then categorizing such points into $Q_s$ corresponds to an optimal partition of $P$ to $Q$.
	
	\label{cor2}
\end{corollary}

\proof {Proof.}
Facts (a) and (b) follow directly from \mbox{Theorem~\ref{the1}} (by contradiction).

(c) Let a point $q_t$ between $q_{s1}$ and $q_{s2}$ be in another group $Q_t$ (associated with $p_t$) of $\mathcal{S^*}$.
By \mbox{Proposition~\ref{pro1}(c)},
$d(p_t, Q_t/q_t) \leq d(p_t, Q_t)$.
Since $q_t$ is weakly dominated by $I_{\{q_{s1}, q_{s2}\}}$,
$p_s' \preceq q_t$ ($p_s'$ being the position for $p_s$ to move to cover $Q_s$).
It follows that $d(p_s, Q_s) = d(p_s, Q_s \bigcup q_t)$.
Therefore,
$d(p_s, Q_s) + d(p_t, Q_t) \geq d(p_s, Q_s \bigcup q_t) + d(p_t, Q_t/q_t)$.
Since $\mathcal{S^*}$ is already an optimal partition,
by \mbox{Theorem~\ref{the1}},
$d(p_s, Q_s) + d(p_t, Q_t) = d(p_s, Q_s \bigcup q_t) + d(p_t, Q_t/q_t)$ and
transferring $q_t$ from $Q_t$ to $Q_s$ is also an optimal partition of $P$ to $Q$.
%\Halmos
\endproof

Fact (a) presents the determinacy of the groups' associated point in an optimal partition
when their cardinality is one.
Fact (b) shows that the optimal substructure holds in this optimal partitioning,
and Fact (c) provides a sufficient condition for points to enter a group
after given two points belonging to this group in an optimal partition.

\section{Calculating DoM in the Biobjective Case}

In this section,
we present a method to calculate the DoM measure in the biobjective case.
Before introducing the calculation procedure,
for convenience we define a term with respect to points' relationship in a set based on the dominance move.

\begin{definition}
	Let $P$ be a set of points,
	and $a \in P$.
	We call a point $b$ ($b\in P, b\neq a$) the \textit{inward neighbor} of $a$ in $P$ (denoted as $n_{P}(a)$)
	if $b$ has the smallest dominance move distance to $a$;
	that is, $b = \argmin_{p\in P/a} d(p,a)$.
	%	In turn,
	%	$a$ is called the \textit{outward neighbor} of $b$.
\end{definition}

Consider two sets of two-dimensional points,
$P = \{p_1, p_2,...,p_{n}\}$ and $Q = \{q_1, q_2,...,q_{l}\}$.
The dominance move of $P$ to $Q$ can be calculated by four steps below.
Step 1 removes the points that do not affect $D(P,Q)$.
Step 2 is the first attempt of constructing an optimal partition of $P$ to $Q$.
Step 3 tests whether it is an optimal partition or not.
If not,
Step 4 recursively merges points of $Q$ to make all points of $Q$ eventually associated with one point in $P$.

\textit{Step} 1. Remove the dominated points in both $P$ and $Q$,
respectively,
and remove the points of $Q$ that are dominated by at least one point in $P$.

It is clear that these points have no effect on the result of $D(P,Q)$ (Fact (d) of \mbox{Proposition~\ref{pro1}}).
So now in the union set of $P$ and $Q$,
there is only one possibility of two points subject to the dominance relation:
a point from $P$ being dominated by a point from $Q$.

\textit{Step} 2. Denote $R=P\bigcup Q$.
First we regard each point of $Q$ in $R$ as a group.
Then for each point of $Q$,
find its inward neighbor in $R$;
that is,
$\forall q\in Q$,
find a point $r \in R$ such that $r = n_{R}(q)$.
If the point $r \in P$,
then merge $r$ into the group of $q$.
For the case that the point $r \in Q$,
if $q$ and $r$ are already in one group,
do nothing;
else merge the two groups of $q$ and $r$ into one group.

Now we can see that each point of $Q$ and its inward neighbor belong to one group.

\textit{Step} 3. If there exists no point $q \in Q$ such that $q = n_{R}(n_{R}(q))$
(i.e., two points are the inward neighbor of each other) in any group,
then the procedure ends;
these groups construct an optimal partition of $P$ to $Q$ (which will be presented in \mbox{Theorem~\ref{the2}} below).
Otherwise,
the procedure continues.

Now,
we show that Step 2 constructs an optimal partition of $P$ to $Q$
provided that $\nexists q \in Q$ satisfying that $q = n_{R}(n_{R}(q))$ in any group.
To prove this,
we first present that it is a partition of $P$ to $Q$
(i.e., in each group there is one and only one point belonging to $P$).

\begin{lemma}
	Let $P$ and $Q$ be two sets of nondominated points in the biobjective case.
	Place each point of $Q$ and its inward neighbor in $P\bigcup Q$ into the same group by Step 2.
	The resulting groups construct a partition of $P$ to $Q$
	provided that $\nexists q \in Q | q = n_{R}(n_{R}(q))$ in any group.
	\label{lemma1} 	
\end{lemma}

\proof {Proof.}
First,
we can see from Step 2 that in any group there is no more than one point of $P$.
Now we only need to present that there is at least one point of $P$ in each group.
Apparently,
if there is no point of $P$ in one group,
some points in this group will form a directed circle
whose edges are the dominance move of their inward neighbor to themselves
(since the number of the points is equivalent to that of the edges).
The case that $q = n_{R}(n_{R}(q))$ can be seen as the circle having only two vertexes.
Next,
we prove (by contradiction) that in any group there exists no such a circle having more than two vertexes.

Assume that there exists one group,
in which some points form a directed circle.
Let these points be $q_1, q_2,..., q_n$
(where $n \geq 3$),
and also suppose that they are sorted with their first objective,
i.e., $q^1_1 < q^1_2 < ... < q^1_n$.
Since they are nondominated to each other,
we have that $q^2_1 > q^2_2 > ... > q^2_n$ for the second objective.
It follows that the inward neighbor of $q_1$ is $q_2$.
This implies that $q_1$ is the inward neighbor of $q_n$
since these $n$ points form a directed circle.
However,
by $q^2_1 - q^2_n > q^2_2 - q^2_n$,
we have that $d(q_1, q_n) > d(q_2, q_n)$.
Hence,
$q_1$ is not the inward neighbor of $q_n$,
a contradiction.
%\Halmos
\endproof

Now we present that this partition is an optimal partition.

\begin{theorem}
	The partition of $P$ to $Q$ formed by grouping each point of $Q$ and its inward neighbor in $P\bigcup Q$
	is an optimal partition of $P$ to $Q$.
	\label{the2}
\end{theorem}

\proof {Proof.} (By contradiction)
Assume that the partition formed by grouping each point of $Q$
and its inward neighbor is not an optimal partition.
Then we can find an optimal partition $\mathcal{S^*}(P,Q)$
in which there exists at least one point $q \in Q$ and its inward neighbor not being in the same group.
Let in $\mathcal{S^*}$ $q$ belong to the group $Q_s$ (associated with $p_s$) and
its inward neighbor $a$ ($a\in P$ or $a\in Q$) belong to the group $Q_t$ (associated with $p_t$).
Next,
according to the relation of points $q$ and $p_s$
(whether $q$ weakly dominates $p_s$ or not),
we consider two situations separately.

For the first situation that $q \preceq p_s$,
by \mbox{Proposition~\ref{pro2}} and $d(a, q) < d(p_s, q)$ (the definition of the inward neighbor),
we have
\[
\begin{split}
d(p_s, Q_s) =~& d(p_s, q \bigcup Q_s/q) \\
=~& d(p_s, q) + d(I_{\{p_s, q\}}, Q_s/q) \\
>~& d(a, q) + d(I_{\{p_s, q\}}, Q_s/q) \\
\end{split}
\]
Since $q \preceq p_s$,
it follows that $I_{\{p_s, q\}} = q$.
By this and Fact (f) of \mbox{Proposition~\ref{pro1}},
the above inequality can be expressed as
\[
\begin{split}
d(p_s, Q_s) >~& d(a, q) + d(I_{\{p_s, q\}}, Q_s/q) \\
=~& d(a, q) + d(q, Q_s/q) \\
\geq~& d(a, q \bigcup Q_s/q) \\
=~& d(a, Q_s) \\
\end{split}
\]
If $a \in P$,
then $a = p_t$ (\mbox{Lemma~\ref{lemma1}}).
We thus have that $d(p_s, Q_s) > d(p_t, Q_s)$,
a contradiction with $\mathcal{S^*}$ being an optimal partition (\mbox{Theorem~\ref{the1}}).
If $a \in Q$,
add $d(p_t, Q_t)$ into both sides of the above inequality.
By \mbox{Proposition~\ref{pro2}} and Fact (f) of \mbox{Proposition~\ref{pro1}},
we have
\[
\begin{split}
d(p_s, Q_s) + d(p_t, Q_t) >~& d(a, Q_s) + d(p_t, Q_t) \\
=~& d(a, Q_s) + d(p_t, Q_t/a) + d(I_{\{p_t, Q_t/a\}}, a) \\
\geq~& d(p_t, Q_t/a) + d(I_{\{p_t, Q_t/a\}}, a \bigcup Q_s) \\
=~& d(p_t, Q_t/a \bigcup a \bigcup Q_s) \\
=~& d(p_t, Q_t \bigcup Q_s) \\
\end{split}
\]
A contradiction follows immediately from \mbox{Theorem~\ref{the1}},
thus completing the proof for the situation that $q \preceq p_s$.

Now let us consider the second situation that $q \npreceq p_s$.
Obviously,
$p_s \npreceq q$
(otherwise $p_s$ is the inner neighbor of $q$).
Without loss of generality,
let $p_s$ be \textit{below} $q$ (i.e., $p_s$ be smaller than $q$ on the second objective).
Since $a$ is the inner neighbor of $q$,
the advantage of $q$ over $p_s$ on the first objective will be greater than $d(a, q)$.
On the other hand,
according to the position of the points of $Q_s$ relative to $q$,
we divide them into three subsets $Q_{sa}$, $Q_{sb}$, and $q$ ($Q_s = Q_{sa} \bigcup Q_{sb} \bigcup q$),
where $Q_{sa}$ consists of the points above $q$ and $Q_{sb}$ below $q$.
By \mbox{Proposition~\ref{pro2}},
we have
\[
\begin{split}
d(p_s, Q_{sb}\bigcup q \bigcup Q_{sa}) =~& d(p_s, Q_{sb}) + d(I_{p_s \bigcup Q_{sb}}, q \bigcup Q_{sa}) \\
=~& d(p_s, Q_{sb}) + d(I_{p_s \bigcup Q_{sb}}, q) + d(I_{p_s \bigcup Q_{sb} \bigcup q}, Q_{sa})  \\
\end{split}
\]
Since both $p_s$ and $Q_{sb}$ are below $q$ and $a$ is the inner neighbor of $q$,
the advantage of $q$ over the ideal point of $p_s$ and $Q_{sb}$ on the first objective is still greater than $d(a, q)$,
namely $d(I_{p_s \bigcup Q_{sb}}, q) > d(a, q)$.
Considering $d(I_{p_s \bigcup Q_{sb} \bigcup q}, Q_{sa})$,
since $Q_{sa}$ is above $q$,
and $q$ is above and nondominated with $p_s$ and $Q_{sb}$,
it follows that $d(I_{p_s \bigcup Q_{sb} \bigcup q}, Q_{sa}) = d(q, Q_{sa})$.
Substituting these in the above equality,
we have
\[
\begin{split}
d(p_s, Q_{sb}\bigcup q \bigcup Q_{sa}) =~& d(p_s, Q_{sb}) + d(I_{p_s \bigcup Q_{sb}}, q) + d(I_{p_s \bigcup Q_{sb} \bigcup q}, Q_{sa}) \\
>~& d(p_s, Q_{sb}) + d(a, q) + d(I_{p_s \bigcup Q_{sb} \bigcup q}, Q_{sa})  \\
=~& d(p_s, Q_{sb}) + d(a, q) + d(q, Q_{sa}) \\
\geq~& d(p_s, Q_{sb}) + d(a, q \bigcup Q_{sa}) \\
\end{split}
\]
If $a \in P$,
then $a = p_t$ and a contradiction directly follows from
$d(p_s, Q_{sb}\bigcup q \bigcup Q_{sa}) > d(p_s, Q_{sb}) + d(p_t, q \bigcup Q_{sa})$.
If $a \in Q$,
add $d(p_t, Q_t)$ into both sides of the inequality.
By \mbox{Proposition~\ref{pro2}} and Fact (f) of \mbox{Proposition~\ref{pro1}},
we have
\[
\begin{split}
d(p_s, Q_{sb}\bigcup q \bigcup Q_{sa}) + d(p_t, Q_t) >~& d(p_s, Q_{sb}) + d(a, q \bigcup Q_{sa}) + d(p_t, Q_t) \\
=~& d(p_s, Q_{sb}) + d(a, q \bigcup Q_{sa}) + d(p_t, Q_t/a) + d(I_{\{p_t, Q_t/a\}}, a) \\
\geq~& d(p_s, Q_{sb}) + d(p_t, Q_t/a) + d(I_{\{p_t, Q_t/a\}}, a \bigcup q \bigcup Q_{sa}) \\
=~& d(p_s, Q_{sb}) + d(p_t, Q_t/a \bigcup a \bigcup q \bigcup Q_{sa}) \\
=~& d(p_s, Q_{sb}) + d(p_t, Q_t \bigcup q \bigcup Q_{sa}) \\
\end{split}
\]
A contradiction (by \mbox{Theorem~\ref{the1}}).
This completes the proof of the second situation.
%\Halmos
\endproof

From the proof of the above claim,
we further have
\begin{corollary}
	In all possible optimal partitions of $P$ to $Q$ in the biobjective case,
	any point of $Q$ and its inward neighbor in $P \bigcup Q$ are in the same group.
	\label{cor3}
\end{corollary}

\textit{Step} 4. For any group that has a circle (i.e., two points being the inward neighbor of each other),
replace these two points by their ideal point.
This leads to a new $Q$ (denoted as $Q'$).
Then find the inward neighbor of such an ideal point in $P \bigcup Q'$ and group them.
Go to Step 3.

This step updates the groups which have no point belonging to $P$.
This update will not affect the inner neighbor of the points of $Q$.

\begin{proposition}
	Let $P$ and $Q$ be two sets of nondominated points in the biobjective case,
	and a pair of points $q_1, q_2 \in Q$ be the inward neighbor of each other in $P \bigcup Q$.
	Let $q^*$ be the ideal point of $q_1, q_2$,
	and $Q' = Q / \{q_1, q_2\} \bigcup q^*$.
	Then,
	replacing $q_1, q_2$ with $q^*$ will keep the inner neighbor of any other points of $Q$ unchanged.
	That is,
	for $q$ ($q\in Q, q \notin \{q_1, q_2\}$),
	if $n_{P\bigcup Q}(q) = q_1$ or $q_2$,
	then $n_{P\bigcup Q'}(q) = q^*$;
	if $n_{P\bigcup Q}(q) = a$ ($a \in P\bigcup Q, a \notin \{q_1, q_2\}$),
	then $n_{P\bigcup Q'}(q) = a$.
	\label{pro3}
\end{proposition}

\proof {Proof.}
To prove this,
we only need to present that for any point $q$ ($q\in Q, q \notin \{q_1, q_2\}$),
$d(q^*, q) = \min\{d(q_1, q), d(q_2, q)\}$.
This implies that if the inner neighbor of $q$ is neither $q_1$ nor $q_2$,
replacing $q_1$ and $q_2$ with $q^*$ will not change its inner neighbor;
if the inner neighbor of $q$ is $q_1$ or $q_2$,
$q^*$ will be its inner neighbor.
Since $q_1$ and $q_2$ are nondominated in the two-dimensional space and
they are the inner neighbor of each other,
there is no point $\in Q$ \textit{between} them.
That is,
some points of $Q$ are above them and the others below them.
Without loss of generality,
let $q_1$ be above $q_2$.
Thus for any point $q_a$ above $q_1$,
we have that $d(q^*, q_a) = d(q_1, q_a)$.
Likewise,
for any point $q_b$ below $q_2$,
$d(q^*, q_b) = d(q_2, q_b)$.
Hence,
for any point $q\in Q/\{q_1, q_2\}$,
$d(q^*, q) = \min\{d(q_1, q), d(q_2, q)\}$.
This completes the proof of the claim.
%%\Halmos
\endproof

The above claim indicates that directly grouping $q^*$ and its inner neighbor
will form a set of groups satisfying that
any point of $Q'$ and its inner neighbor in $P\bigcup Q'$ being in the same group.
Next,
we give the connection of the optimal partitioning of $P$ to $Q'$ and
the optimal partitioning of $P$ to $Q$.

\begin{theorem}
	If $\mathcal{S}$ is an optimal partition of $P$ to $Q'$,
	then $\mathcal{S}$ will be an optimal partition of $P$ to $Q$,
	and also $D(P,Q) = D(P,Q')$.
	\label{the3}
\end{theorem}

\proof {Proof.}
We denote $|\mathcal{S}|$ as the sum of the dominance move distances of the partition $\mathcal{S}$.
Since $\mathcal{S}(P,Q')$ is an optimal partition of $P$ to $Q'$,
$D(P,Q') = |\mathcal{S}(P,Q')|$.
Denote $q^*$ in $Q'$ as the ideal point of $q_1$ and $q_2$ in $Q$.
Apparently,
splitting $q^*$ into $q_1$ and $q_2$ in $\mathcal{S}(P,Q')$ leads to a partition of $P$ to $Q$.
Now we present $\mathcal{S}(P,Q)$ is an optimal partition of $P$ to $Q$.

Assume that $\mathcal{S^*}(P,Q)$ is an optimal partition of $P$ to $Q$,
namely $D(P,Q) = |\mathcal{S^*}(P,Q)|$.
By \mbox{Corollary~\ref{cor3}},
$p_1$ and $p_2$ are in one group,
denoted as $Q_t$ (associated with $p_t$),
in $\mathcal{S^*}(P,Q)$.
Replacing $p_1$ and $p_2$ with their ideal point $q^*$ leads to a partition of $P$ to $Q'$ ($\mathcal{S^*}(P,Q')$).
Since $\mathcal{S}$ is an optimal partition of $P$ to $Q'$,
$|\mathcal{S^*}(P,Q')| \geq |\mathcal{S}(P,Q')|$.
On the other hand,
since $q^*$ is the ideal point of $q_1$ and $q_2$,
by \mbox{Equation~(\ref{eq:DoMpoint})},
we have that $d(p_t, Q_t) = d(p_t, Q_t/\{q_1, q_2\} \bigcup q^*)$.
It follows that $|\mathcal{S^*}(P,Q)| = |\mathcal{S^*}(P,Q')|$.
Likewise,
$|\mathcal{S}(P,Q')| = |\mathcal{S}(P,Q)|$.
Therefore,
\[
D(P,Q) = |\mathcal{S^*}(P,Q)| = |\mathcal{S^*}(P,Q')| \geq |\mathcal{S}(P,Q')| = |\mathcal{S}(P,Q)|
\]
This means that $\mathcal{S}(P,Q)$ is an optimal partition of $P$ to $Q$,
and also $D(P,Q) = D(P,Q')$.
%%\Halmos
\endproof

This theorem,
coupled with \mbox{Theorem~\ref{the2}},
ensures that by recursively removing the circle in newly-generated groups (in order to make them associated with one point of $P$),
the final formed partition corresponds to an optimal partition of $P$ to $Q$.

Next,
we consider the computational complexity of the procedure.
For simplicity,
let both $P$ and $Q$ have the same cardinality $N$ (i.e., $|P| = |Q| = N$).
According to \cite{Bentley1978},
removing the dominated points for the biobjective case (Step 1) requires $O(N\ln N)$ comparisons.
In the biobjective case,
the inner neighbor of the points of $Q$ should be in the range determined by their left and right neighbors in $Q$.
Therefore,
it only needs to check a constant number of points (three on average) to find the inner neighbor of one point of $Q$.
So the operations of Step 2 require $O(N)$ comparisons.
Since at most ($N-1$) circles are generated,
the recursion of Steps 3 and 4 happens at most ($N-1$) times.
Likewise,
finding the inner neighbor of a newly-generated ideal point requires a constant number of comparisons
(also no need to update the inner neighbor of other points of $Q$ according to \mbox{Proposition~\ref{pro3}}).
Thus,
the operations in Steps 3 and 4 require $O(N)$ comparisons.
To summarize,
the overall computational complexity of the procedure is $O(N\ln N)$.

\begin{figure}[tb]
	\begin{center}
		\footnotesize
		\begin{tabular}{@{}c@{}c@{}c@{}}
			\includegraphics[scale=0.35]{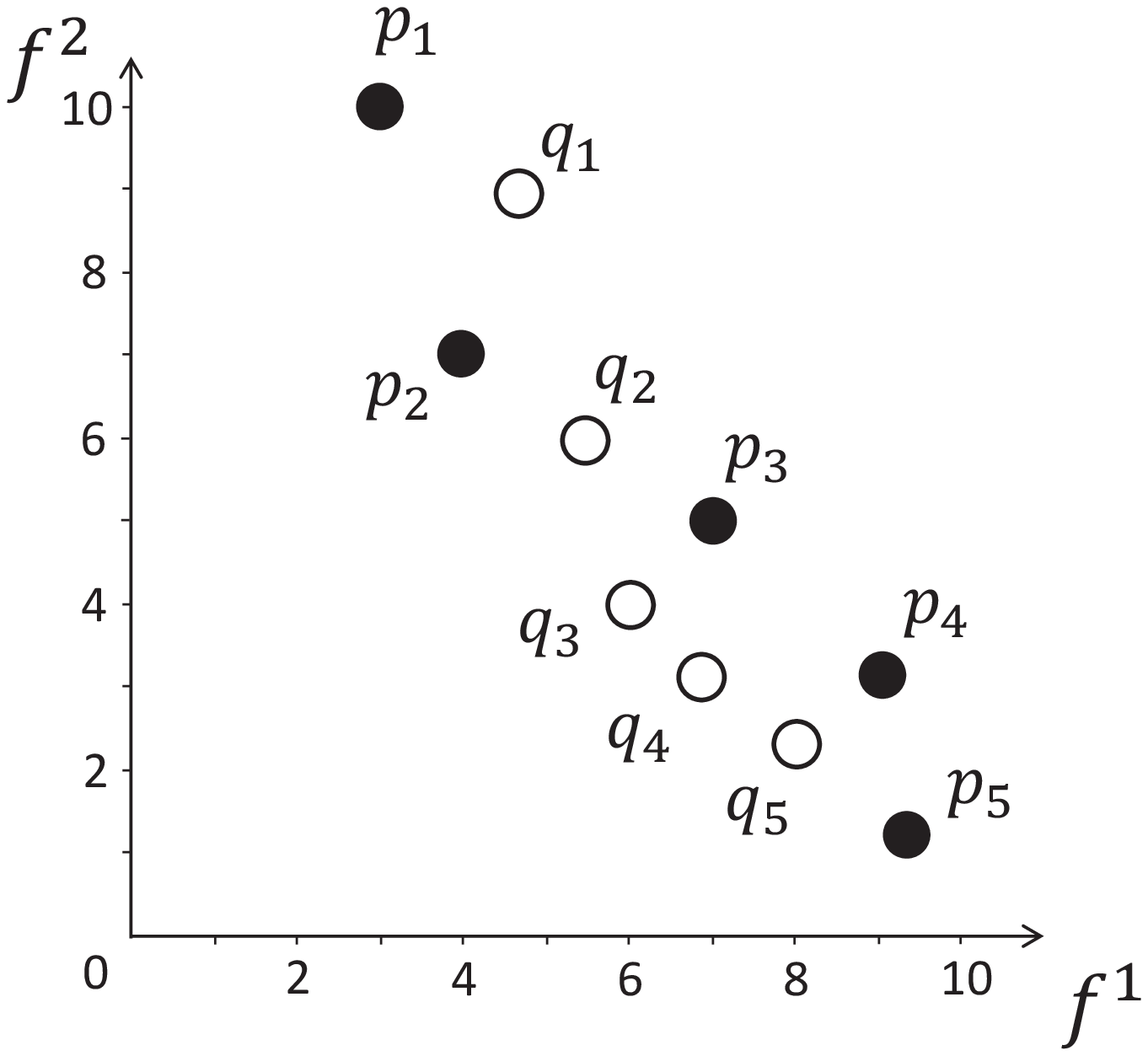}~~ & ~~
			\includegraphics[scale=0.35]{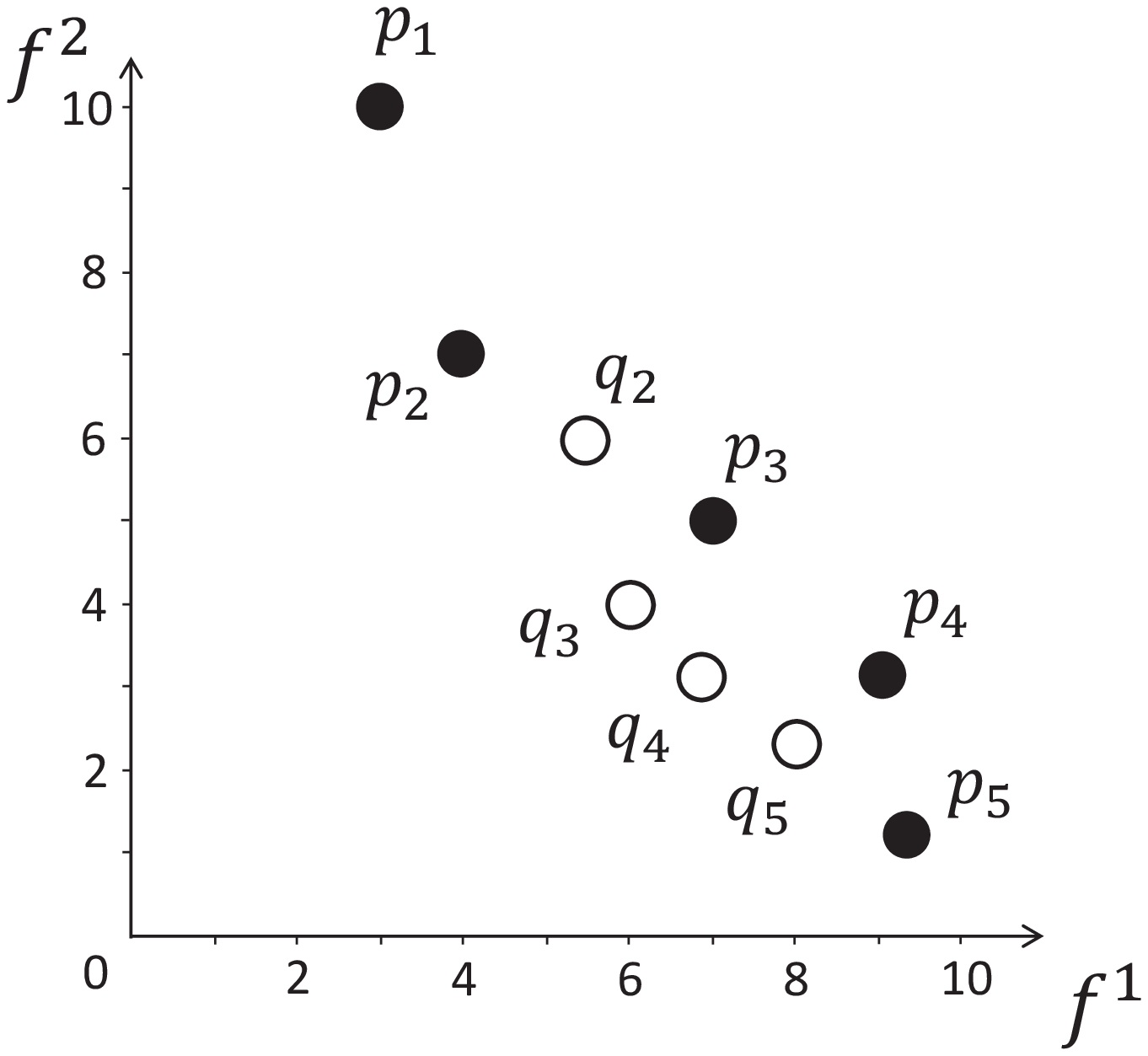}~~ & ~~
			\includegraphics[scale=0.35]{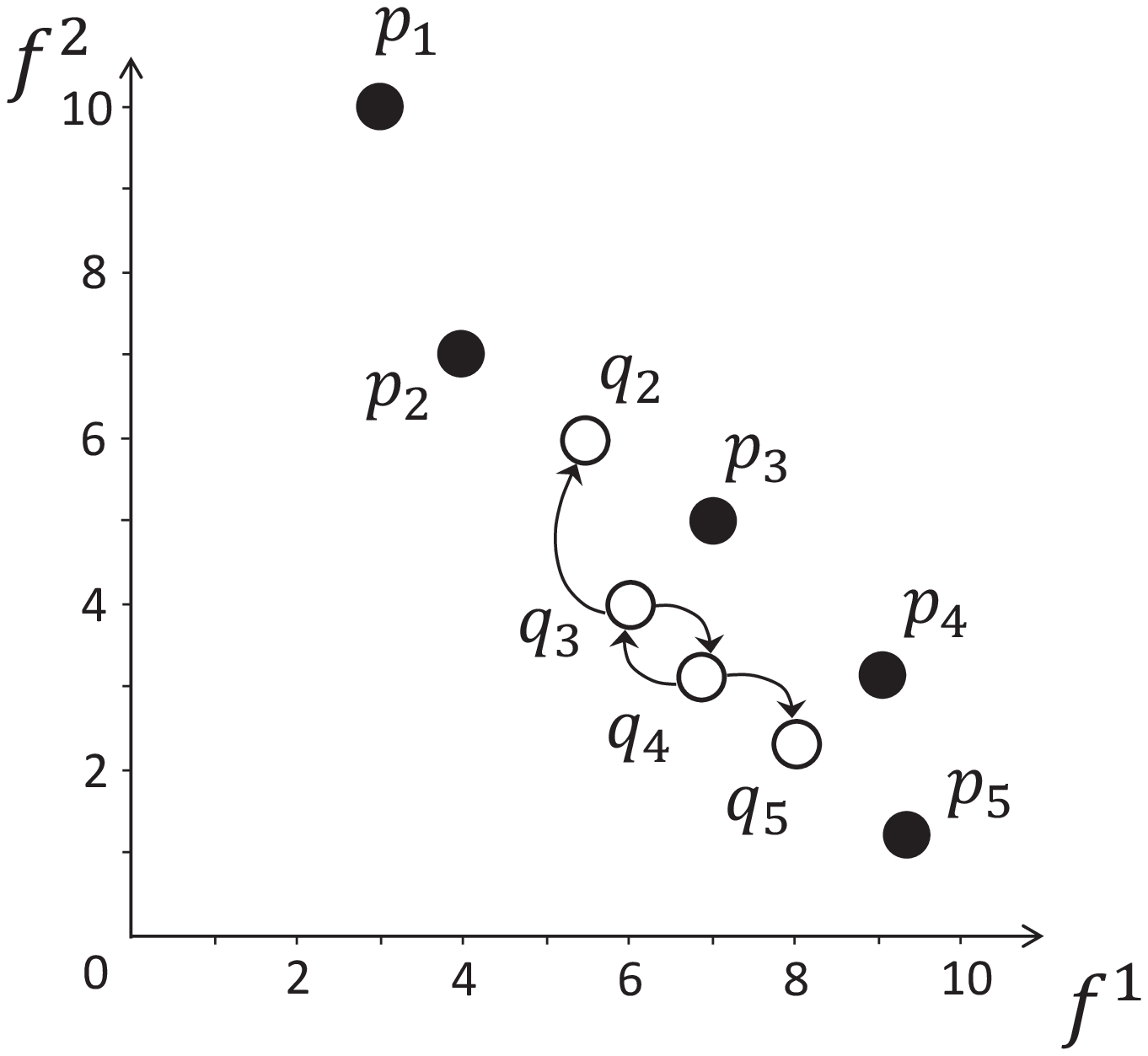} \\
			(a)  & (b)  & (c) \\
			\includegraphics[scale=0.35]{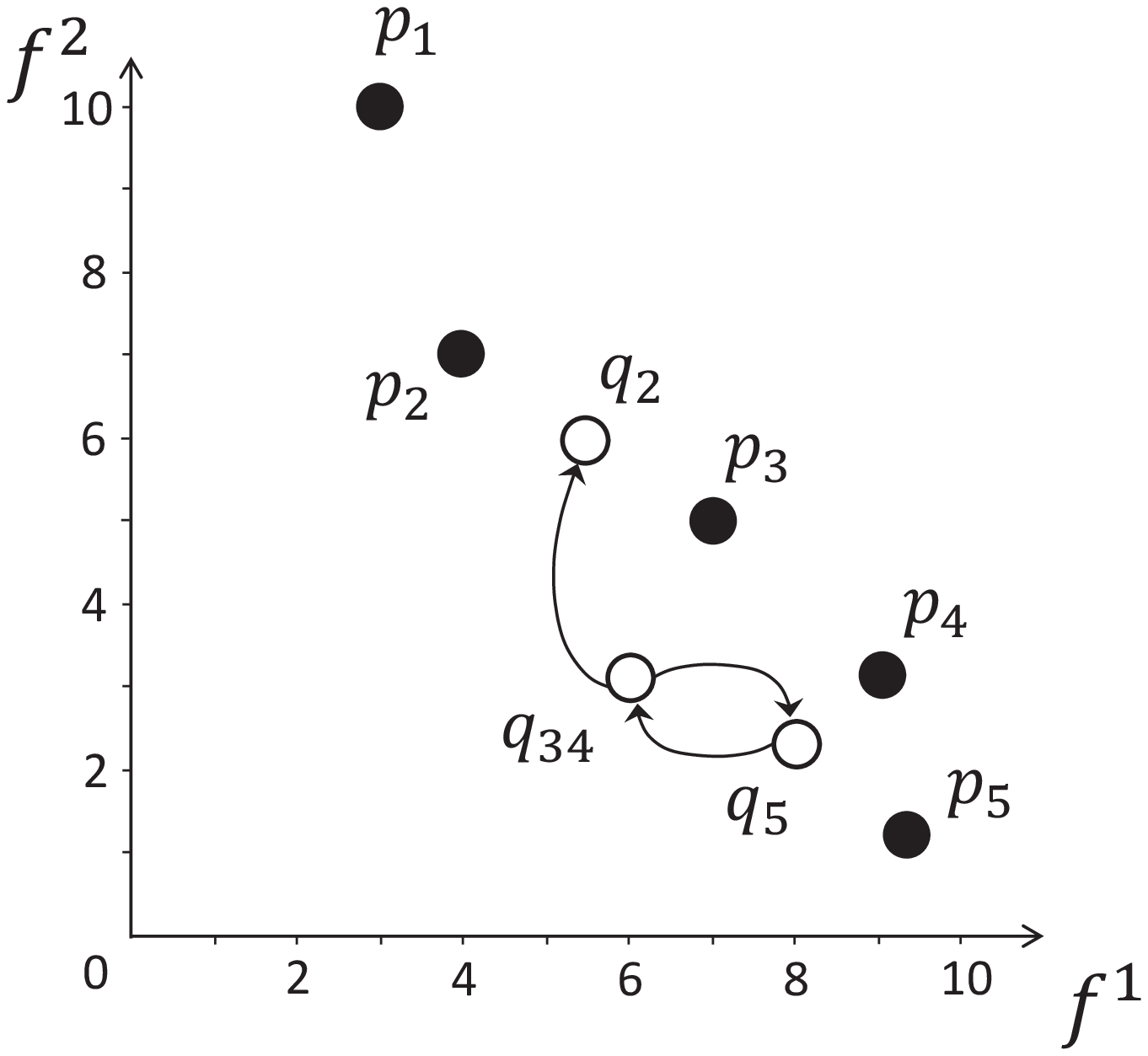}~~ & ~~
			\includegraphics[scale=0.35]{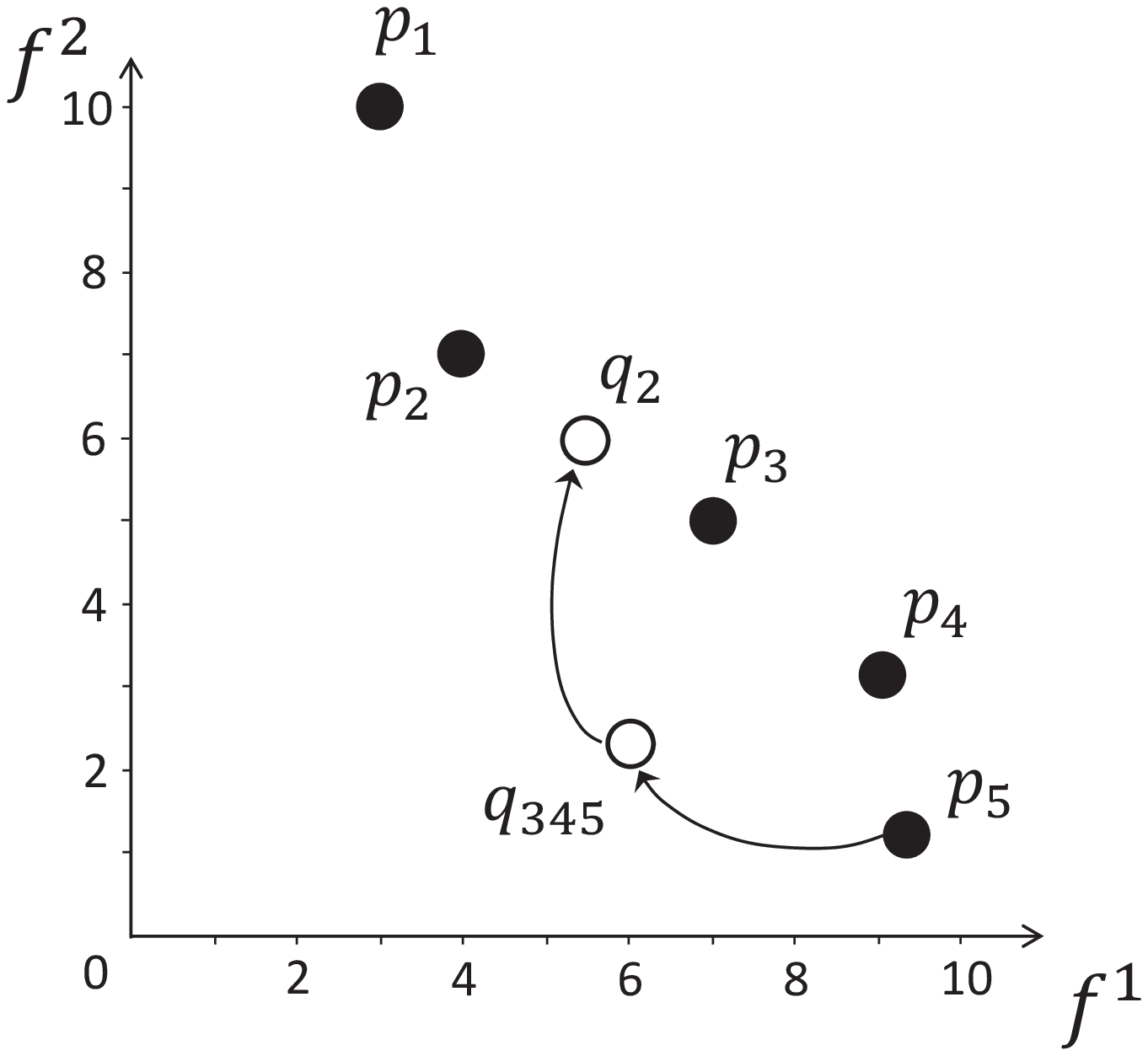}~~ & ~~
			\includegraphics[scale=0.35]{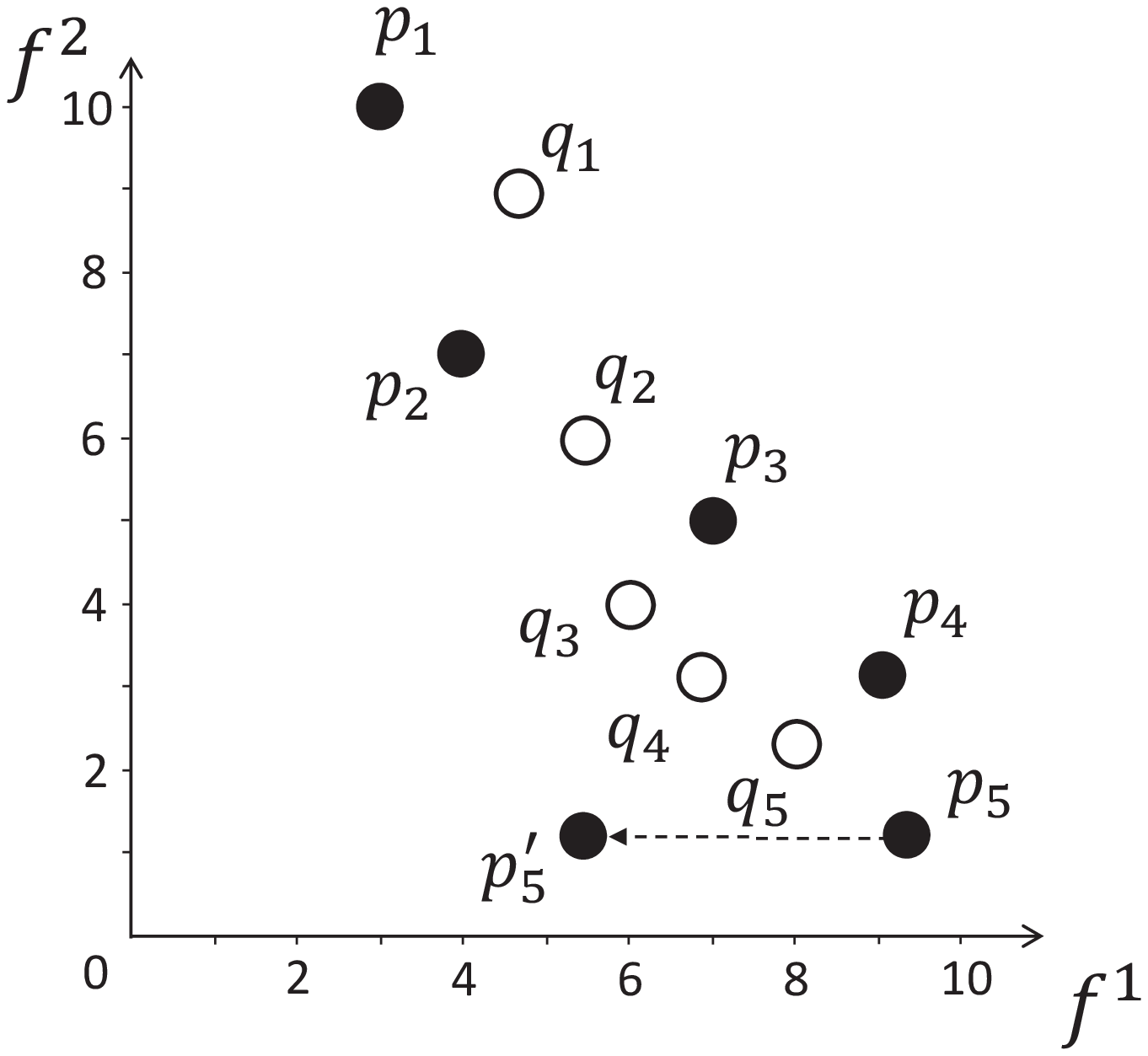} \\
			(d)  & (e)  & (f) \\
		\end{tabular}
	\end{center}
	%\vspace{4mm}
	\caption{An example of the process of calculating the DoM measure.
		(a) The original sets of $P$ and $Q$.
		(b) Removing $q_1$.
		(c) Finding the inner neighbor of points $q_2$ to $q_5$.
		(d) Replacing $q_3$ and $q_4$ by $q_{34}$ and finding its inner neighbor $q_5$.
		(e) Replacing $q_{34}$ and $q_5$ by $q_{345}$ and finding its inner neighbor $p_5$.
		(f) The resultant smallest move of $P$ to dominate $Q$.}
	\label{Fig:Calculationexample}
\end{figure}

To illustrate the proposed method,
\mbox{Figure~\ref{Fig:Calculationexample}} gives an example of calculating the DoM measure stepwise.
The considered sets $P$ and $Q$ respectively have five points,
shown in \mbox{Figure~\ref{Fig:Calculationexample}(a)}.
By Step 1,
$q_1$ has been removed (\mbox{Figure~\ref{Fig:Calculationexample}(b)})
since it is dominated by one point of $P$.
Then,
Step 2 finds the inner neighbor of the points $q_2$ to $q_5$
(which is $q_3, q_4, q_3$ and $q_4$, respectively)
and groups them together.
This is shown in \mbox{Figure~\ref{Fig:Calculationexample}(c)}.
Since $q_3$ and $q_4$ form a directed circle,
replace them by their ideal point $q_{34}$ (\mbox{Figure~\ref{Fig:Calculationexample}(d)}).
And then find the inner neighbor of $q_{34}$: $q_5$.
This however leads to another circle of $q_{34}$ and $q_5$.
Remove this circle by replacing $q_{34}$ and $q_5$ by their ideal point $q_{345}$ (\mbox{Figure~\ref{Fig:Calculationexample}(e)}).
And find the inner neighbor of $q_{345}$: $p_5$.
Now the three points $q_2, q_{345}, p_5$ are in one group.
This results in a partition of $P$ to $Q$ that the points $q_2, q_3, q_4, q_5$ are associated with $p_5$.
Therefore,
the dominance move $D(P,Q)$ is the distance from $p_5$ to $p'_5$
($p_5$ moving to $p'_5$ to dominate $q_2, q_3, q_4, q_5$),
which is shown in \mbox{Figure~\ref{Fig:Calculationexample}(f)}.

%%%% Fig. 4 %%%%

Finally,
it is necessary to note that calculating the DoM measure for more than two objectives may become an intractable task.
The method proposed here cannot apply to such cases directly.
The property that points and their inner neighbor are always in the same group
in an optimal partition (\mbox{Theorem~\ref{the2} and Corollary~\ref{cor3}}) does not hold in general.
For example,
let $P$ and $Q$ be two sets of tri-dimensional points:
$p_1 (2.0,2.0,2.0), p_2 (2.0,2.2,1.5), p_3 (3.0, 1.6, 1.6)$
and $q_1 (2.0,1.2,2.1), q_2 (2.0,2.1,1.0), q_3 (4.0, 1.5, 1.5)$.
The inner neighbor of points $q_1$, $q_2$ and $q_3$ is $p_1$, $p_2$ and $p_3$,
respectively.
This forms a partition of $P$ to $Q$ with the dominance move distance being
$d(p_1,q_1) + d(p_2,q_2) + d(p_3,q_3) = 0.8+0.6+0.2 = 1.6$.
However,
if we directly move $p_2$ to the position $(2.0,1.2,1.0)$ to dominate all the three points of $Q$,
we can obtain a smaller move distance ($1.5$),
which is in fact an optimal partition of $P$ to $Q$.

\section{Experimental Studies}

In this section,
we evaluate the proposed DoM measure.
First,
several groups of artificial test cases are introduced to test the effectiveness of
DoM in reflecting a variety of quality aspects.
Then,
a comparison of DoM with two well-established quality measures,
HV and the $\epsilon$ indicator,
is made.
Finally,
DoM is examined further on two realistic problem instances,
one combinatorial and the other continuous.

\subsection{Artificial Examples}

In general,
the quality of solution sets in multiobjective optimization includes three aspects:
convergence, diversity, and cardinality.
Convergence measures the closeness of a solution set to the Pareto-optimal frontier,
diversity quantifies the distribution of a solution set over the optimal frontier,
and cardinality counts the number of (non-dominated) points in a solution set.
Diversity can be further divided into two sub-aspects \cite{Sayin2000}:
uniformity and extensity.
The former considers the distance between points in a solution set and
the latter measures the range of a solution set covering.
A comprehensive quality measure is expected to be capable of capturing all these aspects.

In this section,
we consider four groups of test cases to evaluate DoM in convergence,
uniformity, coverage and cardinality,
respectively.
We construct these test cases in such a way that it is evident
which solution set is better than the other
in one specific aspect of solution quality.
That is,
in each test case
the difference of the two sets lies in only one aspect;
in the other three aspects they perform equally.

%%%% Fig. 5 %%%%
\begin{figure}[tb]
	\begin{center}
		\footnotesize
		\begin{tabular}{cc}
			\includegraphics[scale=0.25]{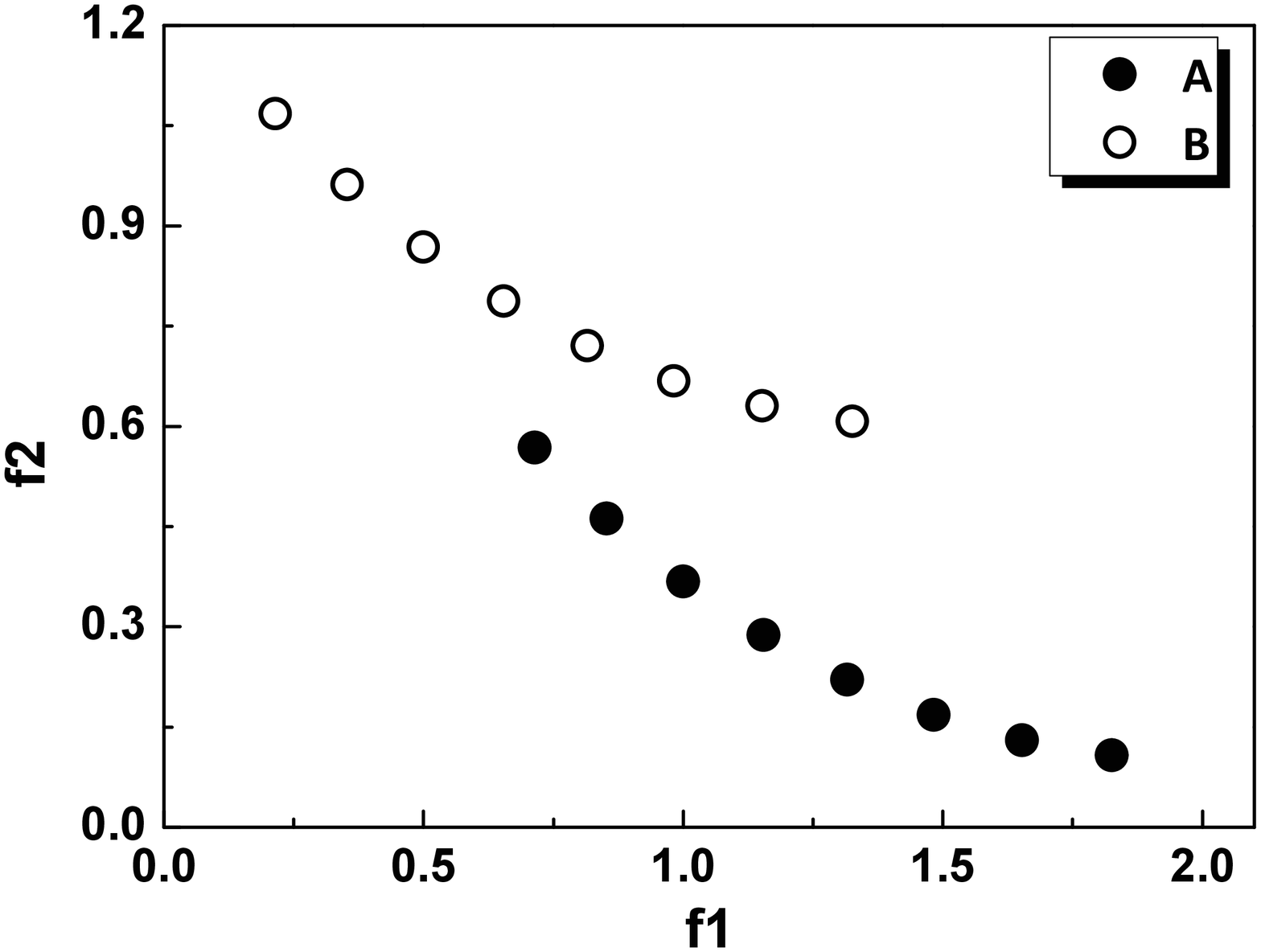} ~~&~~
			\includegraphics[scale=0.25]{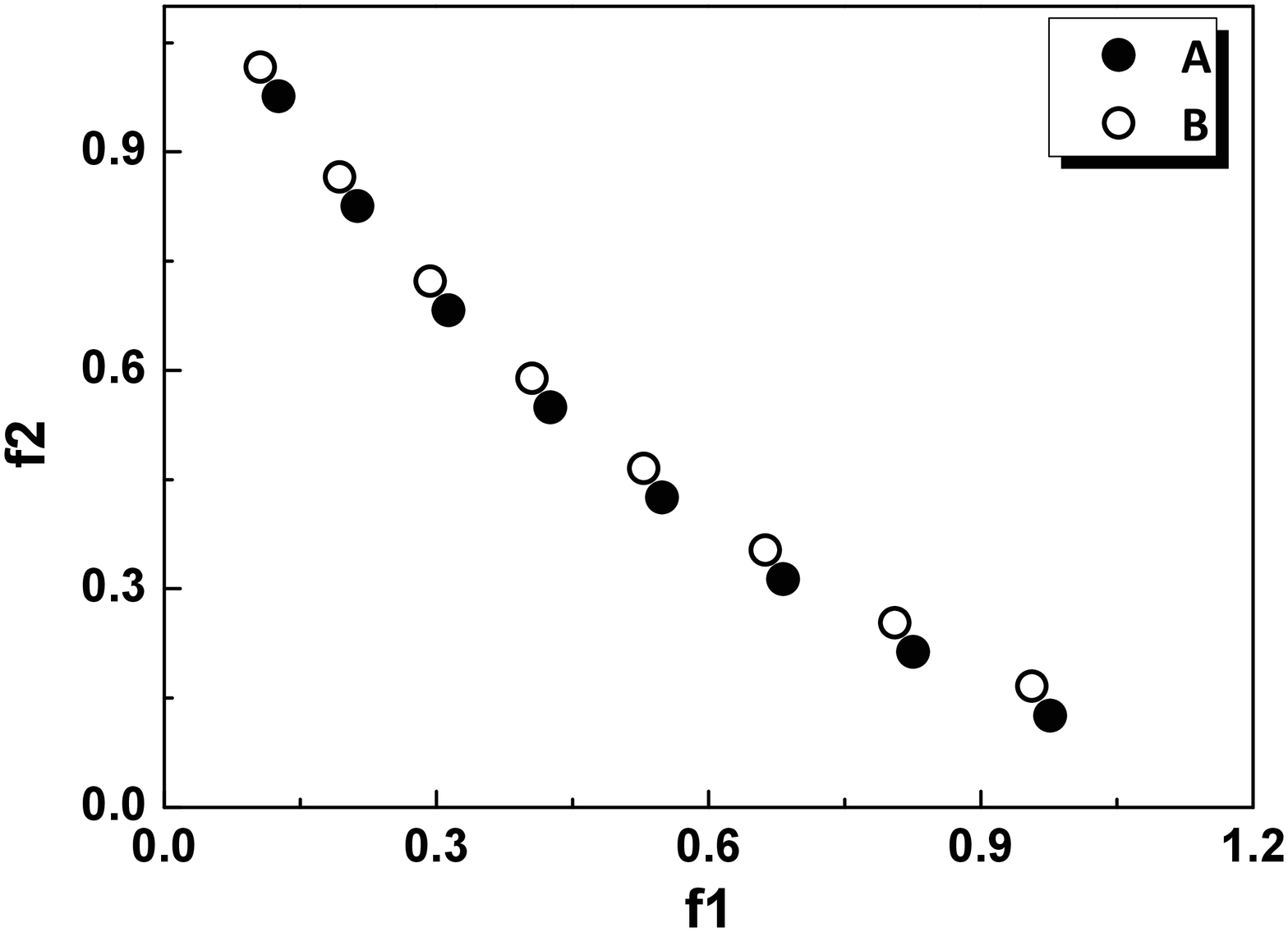} \\
			(a) $D(A,B)=0.500 < D(B,A)=0.680$ ~~&~~ (b) $D(A,B)=0.160 < D(B,A)=0.320$ \\
		\end{tabular}
	\end{center}
	\caption{Convergence test of the DoM measure.
		Each pair of solution sets have the same uniformity, extensity and cardinality.}
	\label{Fig:Convergence}
\end{figure}

Consider convergence first.
Pareto dominance is a central criterion in reflecting the convergence of solution sets.
As we know,
DoM complies with the Pareto dominance criterion.
For two solution sets $A$ and $B$,
if $A$ (weakly) dominates $B$,
$D(A,B) = 0$ (Fact (b) of \mbox{Proposition~\ref{pro1}}).
If $A$ dominates some points of $B$ and $B$ does not dominate any point of $A$,
the DoM measure is likely to prefer $A$ to $B$ (i.e., $D(A,B) < D(B,A)$).
\mbox{Figure~\ref{Fig:Convergence}(a)} is such an example,
where some points of $B$ are dominated by points in $A$ and
the two sets have same quality in the uniformity, extensity and cardinality aspects.
As seen,
the dominance move of $A$ to $B$ is less than that of $B$ to $A$ (0.50 vs 0.68).

Now one may ask what if two solution sets are nondominated completely to each other.
\mbox{Figure~\ref{Fig:Convergence}(b)} is an example to illustrate this situation.
In this example,
we first generate a well-distributed set $A$,
and then generate $B$ through moving $A$ to the upper left a little.
Specifically,
for a point $a_i\in A$,
the corresponding point $b_i \in B$ is generated by
$b^{1}_i = a^{1}_i - 0.02$ and $b^{2}_i = a^{2}_i + 0.04$,
where $b^j_i$ denotes the objective value of point $b_i$ in the $j$th objective.
By this,
any point in the two sets is nondominated,
but $A$ may be seen to have a better convergence than $B$ (in view of the way generating $B$),
which is consistent with the DoM result ($D(A,B)=0.160<D(B,A)=0.320$).
This indicates that the DoM measure prefers the set with better convergence,
even when the two sets are incomparable in terms of the dominance relation of their points.

%%%% Fig. 6 %%%%
\begin{figure}[tb]
	\begin{center}
		\footnotesize
		\begin{tabular}{cc}
			\includegraphics[scale=0.25]{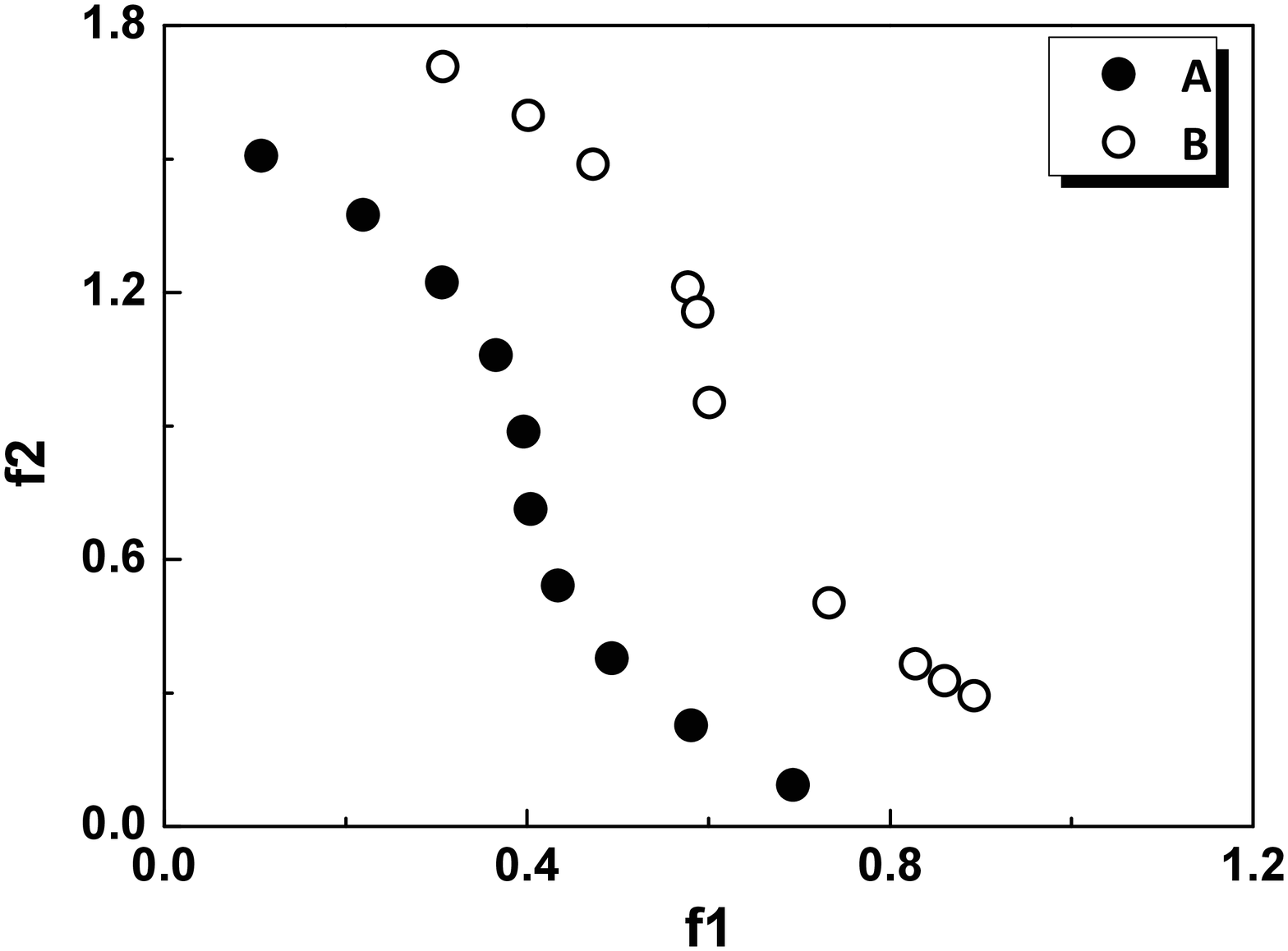} ~~&~~
			\includegraphics[scale=0.25]{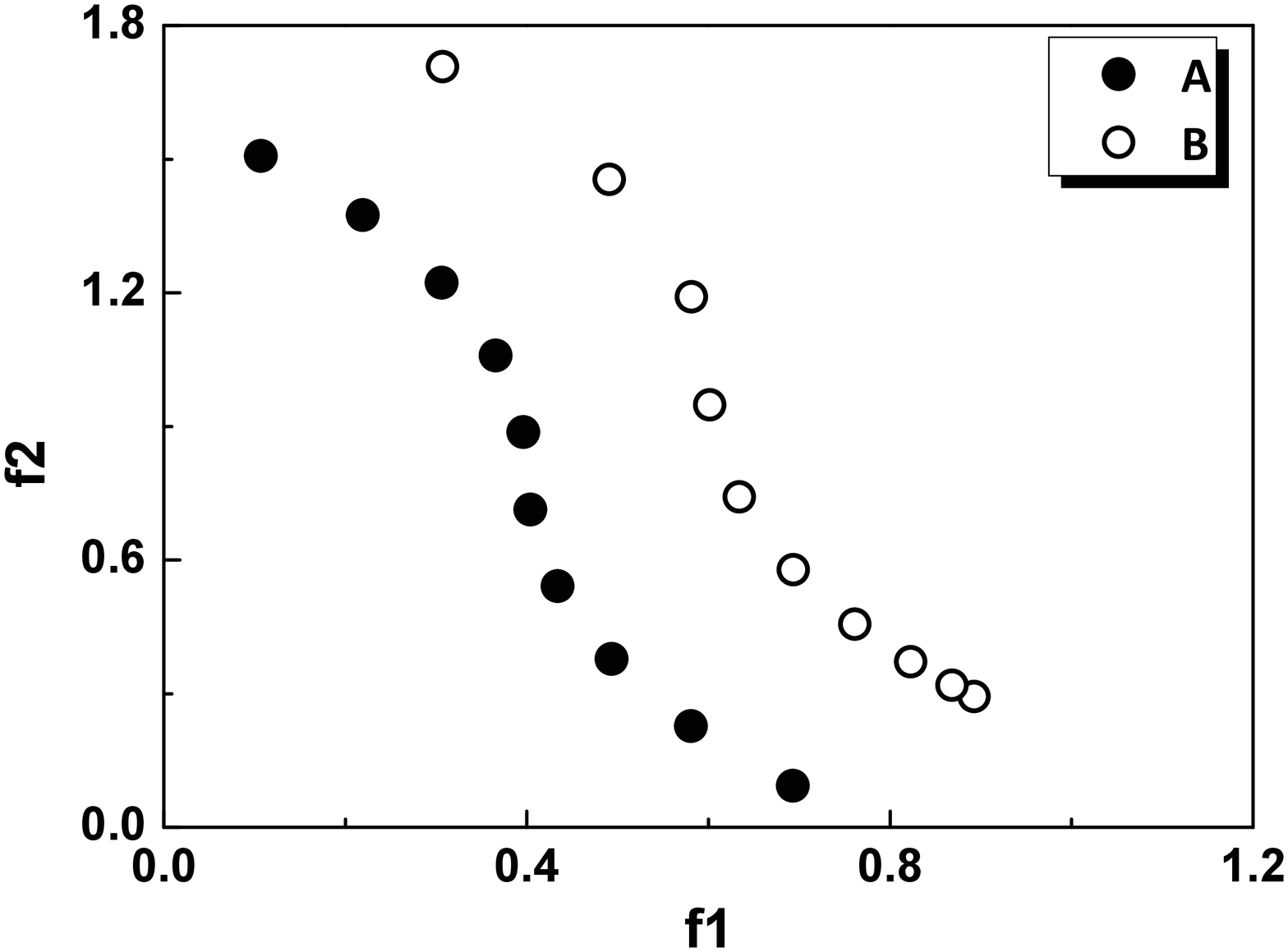} \\
			(a) $D(A,B)=0.186 < D(B,A)=0.277$ ~~&~~ (b) $D(A,B)=0.122 < D(B,A)=0.183$ \\
		\end{tabular}
	\end{center}
	\caption{Uniformity test of the DoM measure.
		Each pair of solution sets have the same convergence, extensity and cardinality.
		For a better observation,
		set $B$ is shifted (added) by $0.2$ on both objectives in the figure.}
	\label{Fig:Uniformity}
\end{figure}

The solution sets in \mbox{Figure~\ref{Fig:Uniformity}} are used to
test DoM in terms of distribution uniformity.
Each pair of sets have the same convergence
(set $B$ being shifted by 0.2 on both objectives in the figure for a better observation),
extensity, and cardinality.
Set $A$ in each pair is distributed uniformly.
Set $B$ in \mbox{Figure~\ref{Fig:Uniformity}(a)} is distributed randomly in the range of set $A$,
and in \mbox{Figure~\ref{Fig:Uniformity}(b)} the distance between neighboring points in set $B$
increases gradually from bottom to top.
As can be seen in the figure,
the evaluation results of DoM indicate its preference for a set of uniformly-distributed points.

%%%% Fig. 7 %%%%
\begin{figure}[tb]
	\begin{center}
		\footnotesize
		\begin{tabular}{cc}
			\includegraphics[scale=0.25]{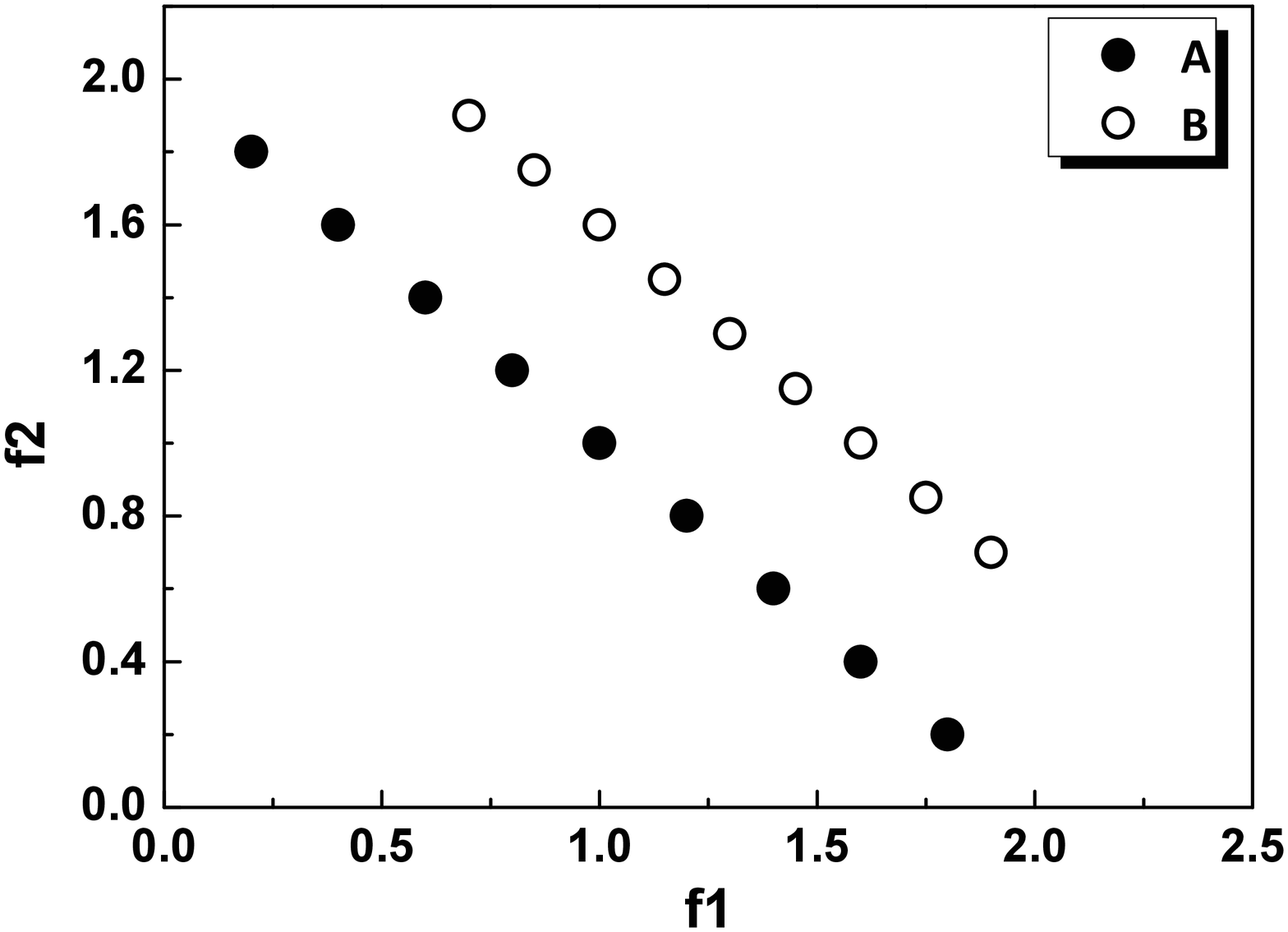} ~~&~~
			\includegraphics[scale=0.25]{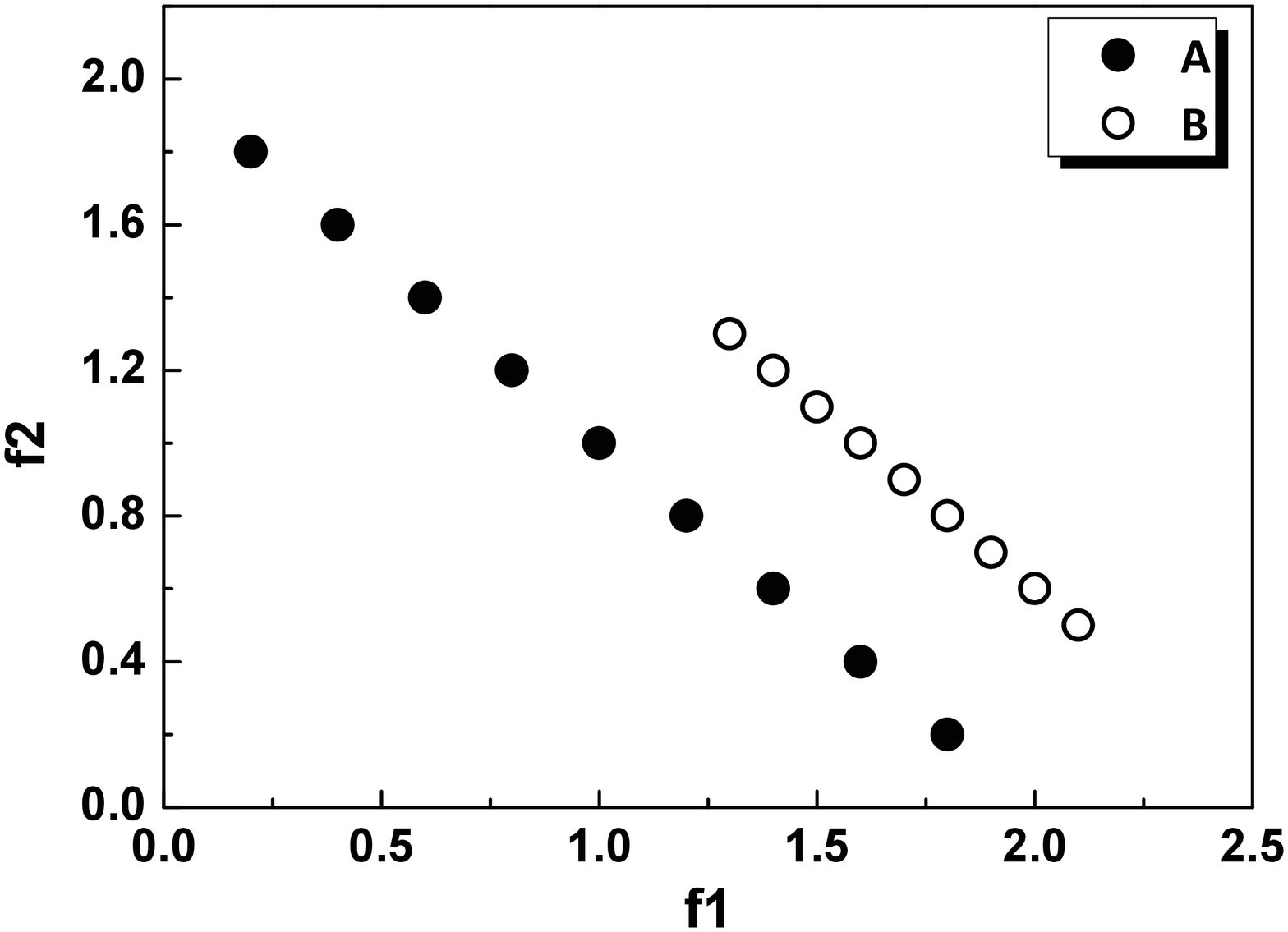} \\
			(a) $D(A,B)=0.400 < D(B,A)=0.600$ ~~&~~ (b) $D(A,B)=0.400 < D(B,A)=0.800$ \\
		\end{tabular}
	\end{center}
	\caption{Extensity test of the DoM measure.
		Each pair of solution sets have the same convergence, uniformity and cardinality.
		For a better observation,
		set $B$ is shifted (added) by $0.3$ on both objectives in the figure.}
	\label{Fig:Extensity}
\end{figure}

The two examples in \mbox{Figure~\ref{Fig:Extensity}} test DoM
in evaluating the distribution extensity of solution sets.
In both examples,
set $A$ is a well-distributed set,
with the range of $1.6$ on both $f_1$ and $f_2$ objectives.
In \mbox{Figure~\ref{Fig:Extensity}(a)},
set $B$ is generated by shrinking $A$ a little,
resulting in its range being $1.2$ on both objectives.
In \mbox{Figure~\ref{Fig:Extensity}(b)},
set $B$ is distributed uniformly in the range of the five bottom right points of $A$,
and thus its range is $0.8$.
As shown,
DoM is able to accurately reflect the extensity of solution sets ---
a set with better extensity has a smaller dominance move to its competitor.

%%%% Fig. 8 %%%%
\begin{figure}[tb]
	\begin{center}
		\footnotesize
		\begin{tabular}{cc}
			\includegraphics[scale=0.25]{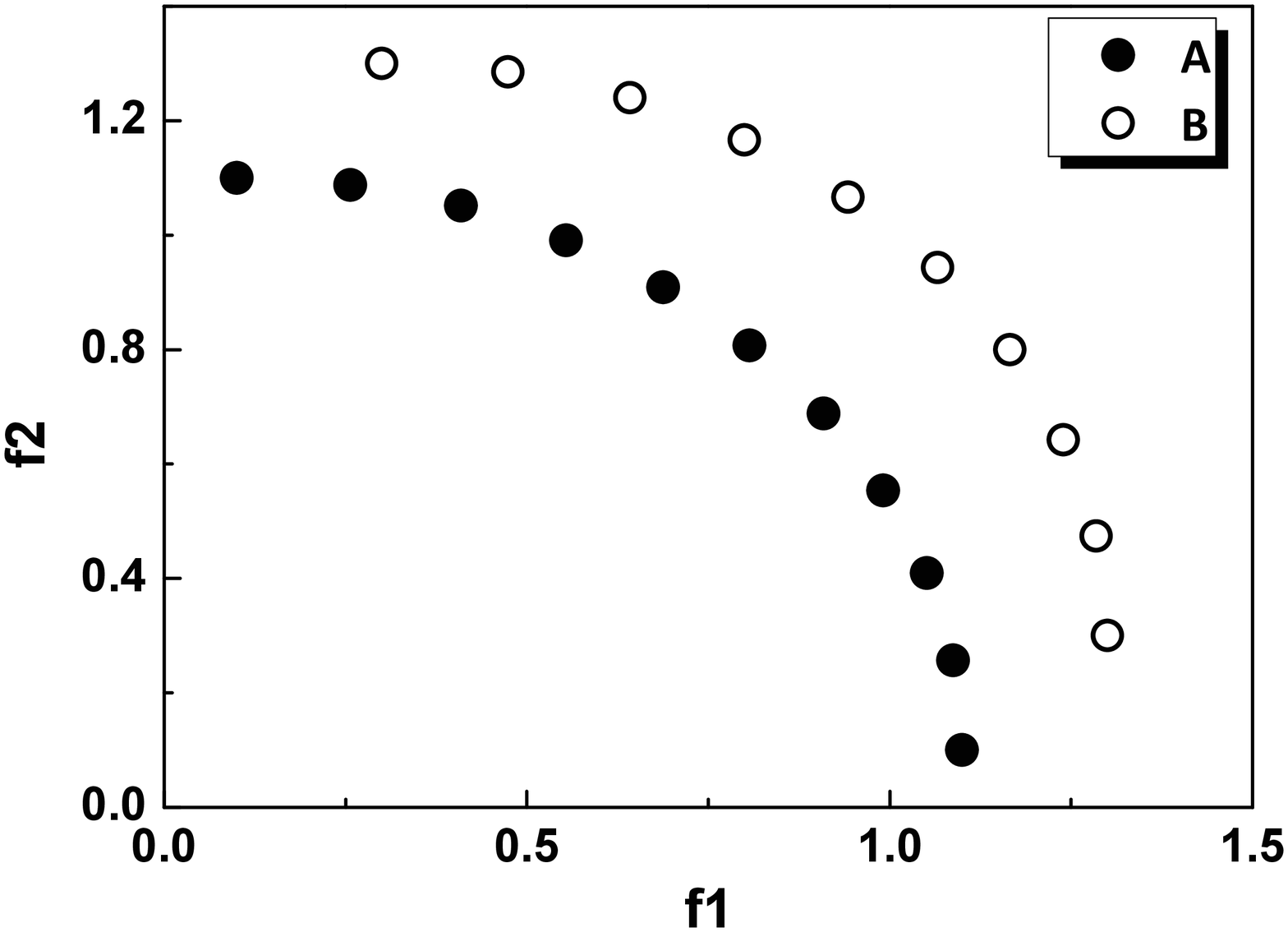} ~~&~~
			\includegraphics[scale=0.25]{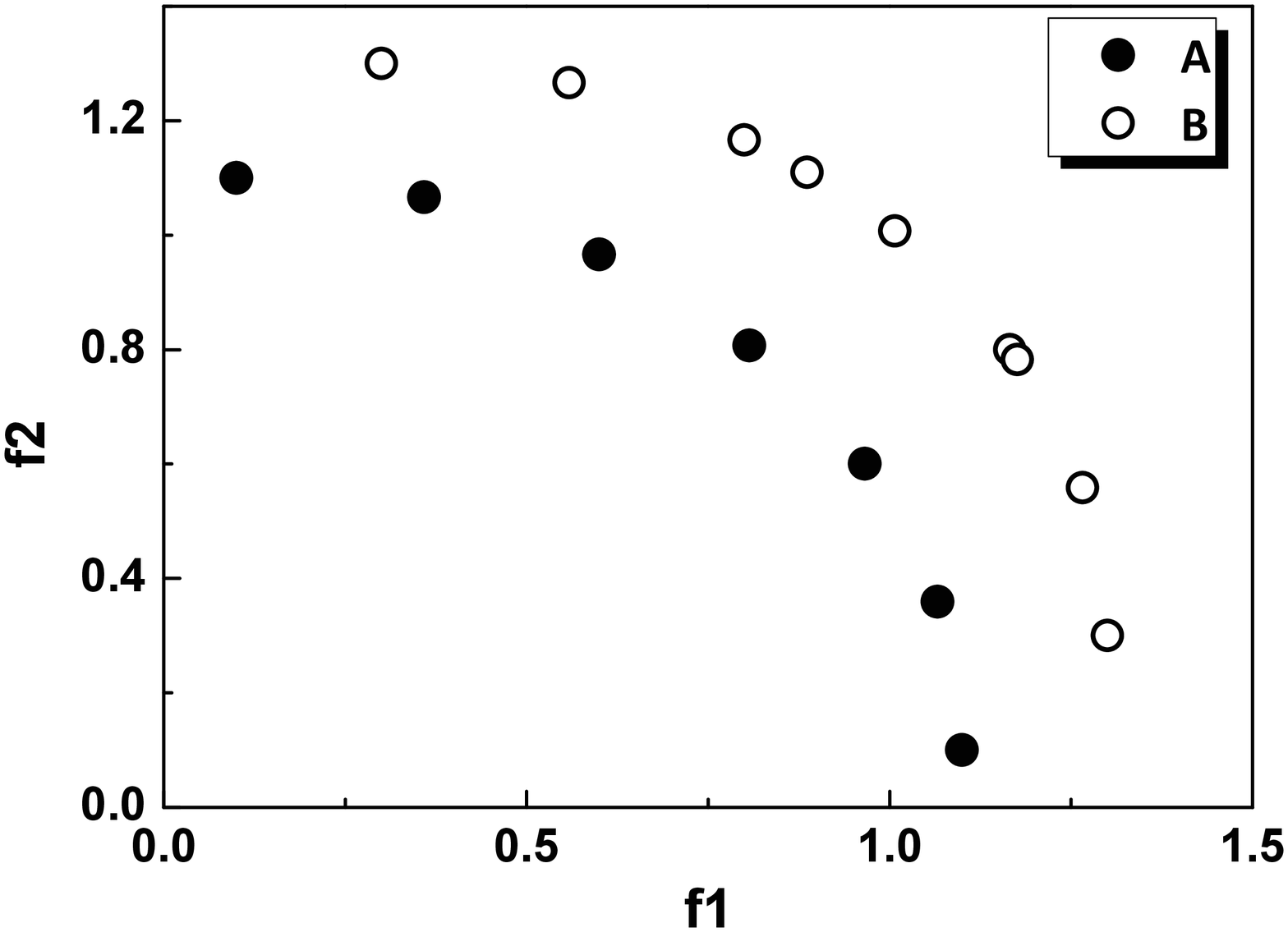} \\
			(a) $D(A,B)=0.164 < D(B,A)=0.352$ ~~&~~ (b) $D(A,B)=0.074 > D(B,A)=0.000$ \\
		\end{tabular}
	\end{center}
	\caption{Cardinality test of the DoM measure.
		The pair of solution sets in (a) have the same convergence, uniformity and extensity,
		but $|A| = 11$ and $|B| = 10$.
		In (b), set $B$ is generated by adding two points to $A$,
		and so $|A| = 7$ and $|B| = 9$.
		For a better observation,
		set $B$ is shifted (added) by $0.2$ on both objectives in the figure.}
	\label{Fig:Cardinality}
\end{figure}

\mbox{Figure~\ref{Fig:Cardinality}} verifies the DoM measure in evaluating the cardinality of solution sets.
In \mbox{Figure~\ref{Fig:Cardinality}(a)},
the two sets perform equally in terms of convergence, uniformity and extensity,
but set $A$ has one more point than $B$,
thus being preferred by DoM ($D(A,B)=0.164<D(B,A)=0.352$).
\mbox{Figure~\ref{Fig:Cardinality}(b)} is an interesting case,
where set $B$ is generated by adding two new points to $A$
which is a set of uniformly-distributed points.
This causes that $B$ has a worse uniformity than $A$.
However,
$B$ provides more information to the decision maker than $A$,
and should be considered better.
In fact,
$B$ weakly dominates $A$ but $A$ does not.
The DoM result can reflect this information:
$D(A,B)=0.074>D(B,A)=0$.

\subsection{Comparison with Other Quality Measures}

In this section,
we compare DoM with two popular quality measures in multiobjective optimization,
the $\epsilon$ indicator and HV.
As described in Section 2.2,
these two measures have some desirable features and
both aim to provide a comprehensive evaluation of solution sets' quality.

In this comparison,
we consider the solution set examples from \mbox{Figures~\ref{Fig:Epsilon}} and \ref{Fig:HV} in Section 2.2,
where the $\epsilon$ indicator and HV fail to accurately compare the sets,
respectively.
This allows us to see if the proposed measure is able to provide a reliable result on the examples
where its peers struggle.

\mbox{Table~\ref{Table:comparison}} shows evaluation results of the three measures.
For the example in \mbox{Figure~\ref{Fig:Epsilon}},
set $P$ has a better distribution and more points than set $Q$,
but the $\epsilon$ indicator cannot distinguish between them ($I_{\epsilon}(P, Q) = I_{\epsilon}(Q, P) = 1.0$).
In contrast,
the HV and DoM measures can accurately reflect the difference of the two sets,
with $P$ having a better evaluation result always.

%%%% Table I %%%%
\begin{table}[tbp]
	\begin{center}
		\begin{scriptsize}
			\caption{Evaluation results of the DoM, $\epsilon$ and hypervolume (HV) measures on the solution set examples in Figures~\ref{Fig:Epsilon} and \ref{Fig:HV}.
				For the hypervolume, the reference point $(10,10)$ is used in Figure~\ref{Fig:Epsilon}'s example,
				and $(10, 11)$ in Figure~\ref{Fig:HV}'s example.
				A better result is highlighted in boldface.}
			\vspace{2pt}
			\label{Table:comparison}
			\begin{tabular}{c|c|c|c} \hline
				Two sets & Hypervolume & $\epsilon$ indicator & DoM \\ \hline
				$P$ vs $Q$ in Figure~\ref{Fig:Epsilon} & \textbf{52} vs 47 & 1.0 vs 1.0 & \textbf{1.0} vs 4.0\\
				\hline
				$P$ vs $Q$ in Figure~\ref{Fig:HV} & 48 vs \textbf{49} & 2.0 vs 2.0 & \textbf{2.0} vs 4.0 \\ \hline
			\end{tabular}%
		\end{scriptsize}
	\end{center}
	\vspace{-2mm}
\end{table}

For the example in \mbox{Figure~\ref{Fig:HV}},
two points in set $Q$ are dominated by set $P$.
Clearly,
$P$ provides more information than $Q$ and is likely to be preferred by the decision maker.
However,
the HV evaluation result on these two sets depends on the choice of the reference point.
If the reference point is set to $(10, 11)$,
set $Q$ will be preferred by HV to $P$.
When the two sets are compared using the $\epsilon$ indicator,
the evaluation results are identical ($2.0$).
This is because the $\epsilon$ indicator only considers the difference on one particular objective
of one particular point of either solution set.
In contrast,
DoM can reflect the advantage of $P$ to $Q$,
with the dominance move of $P$ to $Q$ being less than that of $Q$ to $P$
($D(P,Q)=2.0<D(Q,P)=4.0$).
This indicates that the proposed DoM measure works on some examples on which its peers do not.

\subsection{Real Examples}

In this section,
we further evaluate the DoM measure by considering two pairs of solution sets (obtained by metaheuristics)
on well-established combinatorial and continuous optimization problems,
multiobjective 0--1 knapsack problem \cite{Zitzler1999, Phelps2003} and ZDT \cite{Zitzler2000}.
The metaheuristics used here are two of the most popular algorithms
in the evolutionary multiobjective optimization area,
NSGA-II \cite{Deb2002} and MOEA/D \cite{Zhang2007}.

Each solution set was obtained by a single run of an algorithm.
In the two algorithms,
the population size was set to 100,
and the termination criterion was 100,000 evaluations for the knapsack problem and
30,000 evaluations for ZDT3.
A crossover probability $p_c=1.0$ and a mutation probability $p_m=1/n$ (where $n$ is the
number of decision variables) were used.
For the knapsack problem,
operators for crossover and mutation were the uniform crossover and bit-flip mutation.
For continuous problems,
operators for crossover and mutation were SBX crossover and polynomial mutation \cite{Deb2001}
with both distribution indexes set to 20.

%%%% Fig. 9 %%%%
\begin{figure}[tb]
	\begin{center}
		\footnotesize
		\begin{tabular}{cc}
			\includegraphics[scale=0.25]{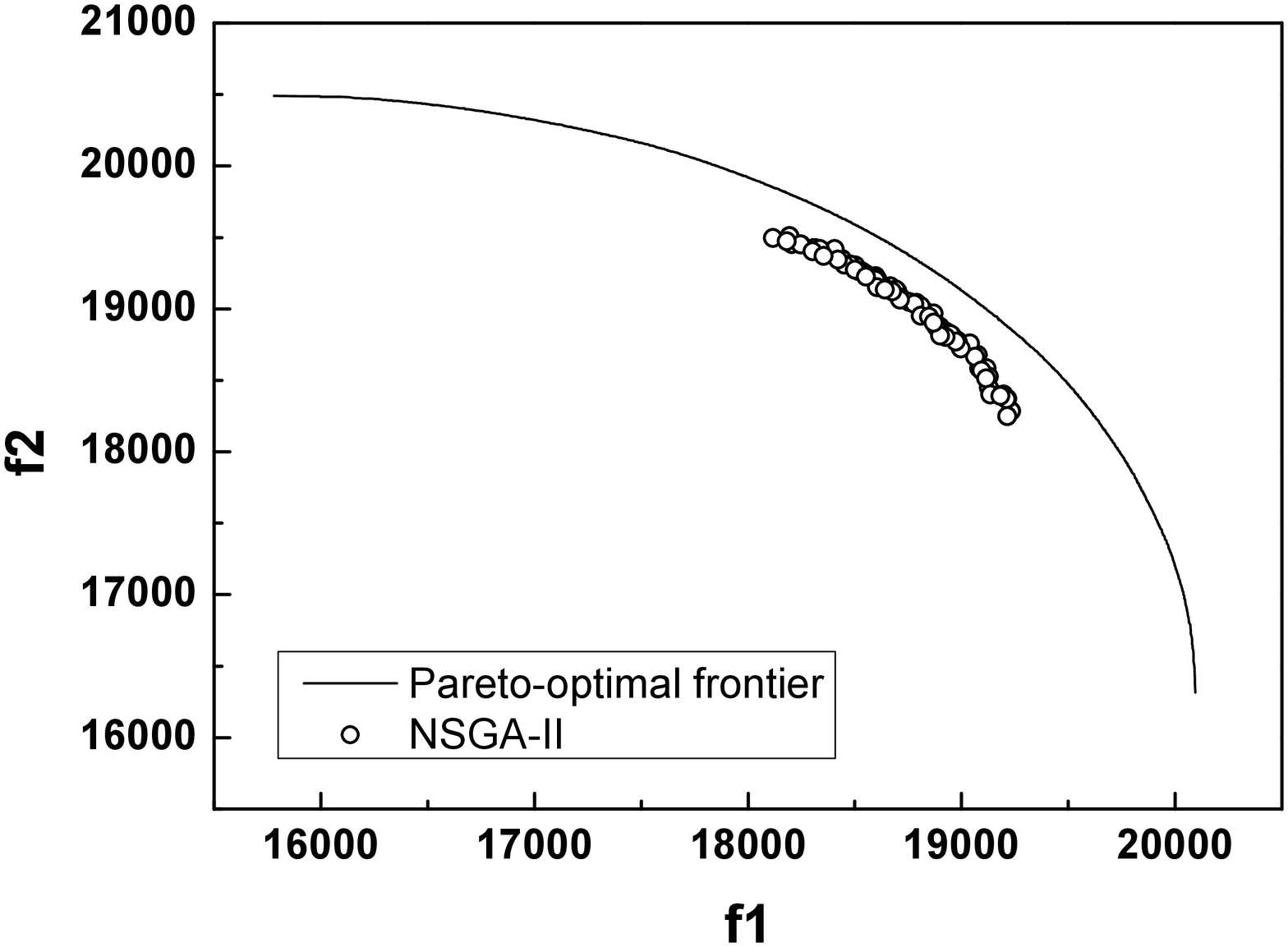} ~~~&~~~
			\includegraphics[scale=0.25]{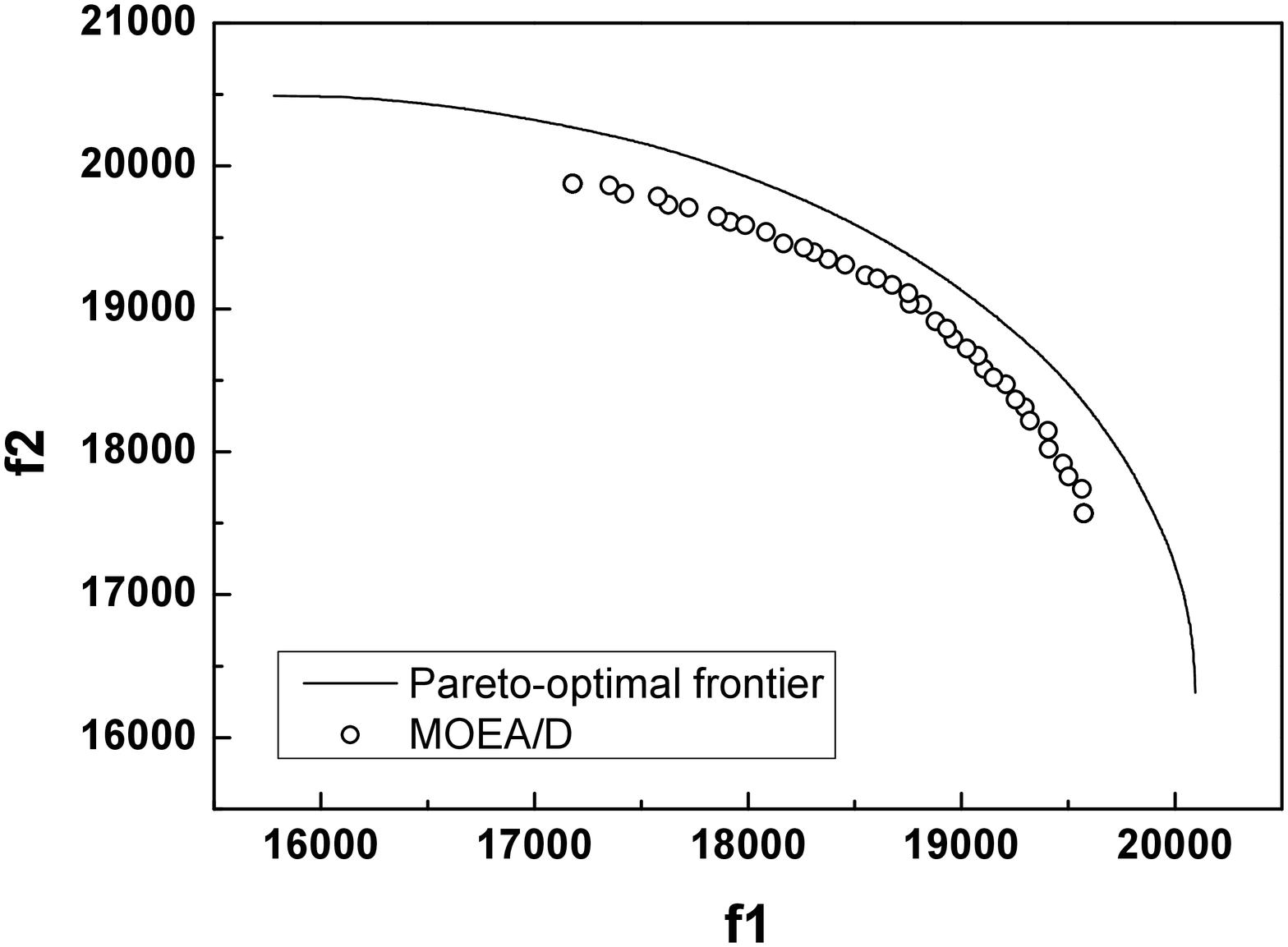} \\
			(a) $D(P_{NSGA-II},P_{MOEA/D})=1244$~~&~~ (b) $D(P_{MOEA/D},P_{NSGA-II})=285$\\
		\end{tabular}
	\end{center}
	\caption{Test of the DoM measure on two solution sets obtained by NSGA-II and MOEA/D on the biobjective 0--1 knapsack problem.}
	\label{Fig:Knapsack}
\end{figure}

%%%% Fig. 10 %%%%
\begin{figure}[tb]
	\begin{center}
		\footnotesize
		\begin{tabular}{cc}
			\includegraphics[scale=0.25]{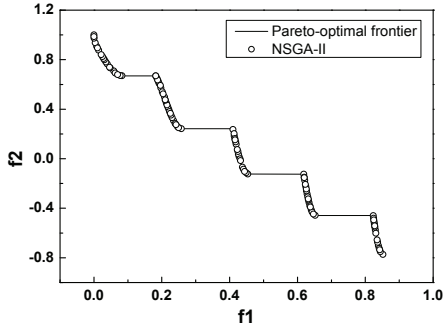} ~~~&~~~
			\includegraphics[scale=0.25]{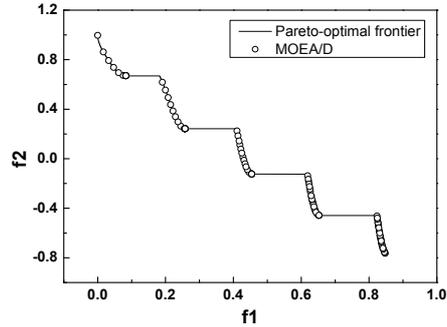} \\
			(a) $D(P_{NSGA-II},P_{MOEA/D})=0.082$ ~~&~~ (b) $D(P_{MOEA/D},P_{NSGA-II})=0.163$  \\
		\end{tabular}
	\end{center}
	\caption{Test of the DoM measure on two solution sets obtained by NSGA-II and MOEA/D on the continuous problem ZDT3.}
	\label{Fig:ZDT3}
\end{figure}

\mbox{Figure~\ref{Fig:Knapsack}} shows two solution sets obtained by NSGA-II and MOEA/D
on the biobjective 0--1 knapsack problem.
The Pareto-optimal frontier of the problem was also plotted for reference.
As can be seen from the figure,
the two sets have similar quality in convergence,
but the set obtained by MOEA/D has clearly better diversity.
This is consistent with the evaluation result of DoM
that the set obtained by MOEA/D is preferred a lot ($D(P_{NSGA-II},P_{MOEA/D})=1244 > D(P_{MOEA/D},P_{NSGA-II})=285$).

The ZDT problem has a discontinuous Pareto-optimal frontier,
to which both algorithms are able to converge,
as shown in \mbox{Figure~\ref{Fig:ZDT3}}.
However,
in contrast to the solution set of NSGA-II which has a good coverage over the whole Pareto-optimal frontier,
more solutions of MOEA/D concentrate in the lower right part of the optimal frontier.
DoM can reflect this quality difference,
and it indicates a preference for the set of NSGA-II over the set of MOEA/D,
as expected.

\section{Conclusions}

This paper proposes a quality measure,
dominance move (DoM),
to compare a pair of solution sets in multiobjective optimization.
DoM measures the required minimum move for one set to weakly Pareto dominate the other.
The proposed measure is intuitive,
and it is a natural reflection of the difference between two solution sets in the context of multiobjective optimization.
DoM can be of high practicability given its desirable properties,
such as a comprehensive coverage of solution set quality,
compliance with Pareto dominance,
and no need of any problem knowledge and parameters.

Systematic experiments have been conducted to evaluate DoM in terms of the convergence,
uniformity, extensity, and cardinality of solution sets on four groups of artificial test cases.
A comparison with two popular quality measures has been made to demonstrate the
strength of DoM.
Real test cases have been considered by testing DoM on two pairs of solution sets
(obtained by metaheuristics) on combinatorial and continuous optimization instances.
The evaluation results have confirmed the effectiveness of the proposed measure.

An efficient method to calculate the DoM in the biobjective case has been presented.
However,
this method may not be extended directly to the case with more objectives.
An important property in the biobjective case is
that solutions and their inner neighbor are always in the same group in an optimal partition.
This does not hold in general in a higher-dimensional case.
Therefore,
how to efficiently calculate the DoM in the case with three or more objectives
remains to be explored.

%\bibliographystyle{IEEEtran}
%\bibliography{IEEEabrv,bibfile}


\end{document}